\documentclass{article}

\usepackage{microtype}
\usepackage{graphicx}
\usepackage{booktabs} 
\usepackage{multirow}
\usepackage{multicol}
\usepackage{float}
\usepackage{subfig}
\usepackage{paralist, tabularx}
\usepackage{newcommand}
\usepackage{adjustbox}
\usepackage{amssymb,amsmath,bm,color}
\usepackage{mathtools}
\usepackage{pifont}
\newcommand{\bdelta}{{\bm{\delta}}}
\usepackage{hyperref}

\newcommand{\xmark}{\ding{55}}%

\newtheorem{theorem}{Theorem}[section]

\newtheorem{definition}[theorem]{Definition}

\usepackage[accepted]{icml2020}

\icmltitlerunning{CAT: Customized Adversarial Training for Improved Robustness}

\begin{document}

\twocolumn[
\icmltitle{CAT: Customized Adversarial Training for Improved Robustness}



\icmlsetsymbol{equal}{*}

\begin{icmlauthorlist}
\icmlauthor{Minhao Cheng}{to}
\icmlauthor{Qi Lei}{goo}
\icmlauthor{Pin-Yu Chen}{ed}
\icmlauthor{Inderjit Dhillon}{goo}
\icmlauthor{Cho-Jui Hsieh}{to}
\end{icmlauthorlist}

\icmlaffiliation{to}{Department of Computer Science, University of California, Los Angeles, USA}
\icmlaffiliation{goo}{Department of Computer Science, University of Texas, Austin, USA}
\icmlaffiliation{ed}{IBM research AI, Yorktown Heights, USA}

\icmlcorrespondingauthor{Minhao Cheng}{mhcheng@ucla.edu}

\icmlkeywords{Machine Learning, ICML}

\vskip 0.3in
]



\printAffiliationsAndNotice{}  

\begin{abstract}
Adversarial training has become one of the most effective methods for improving robustness of neural networks. However, it often suffers from poor generalization on both clean and perturbed data. 
In this paper, we propose a new algorithm, named Customized Adversarial Training (CAT), which adaptively customizes the perturbation level and the corresponding label for each training sample in adversarial training. We show that the proposed algorithm achieves better clean and robust accuracy than previous adversarial training methods through extensive experiments. 
\end{abstract}

\section{Introduction}
Deep neural networks (DNNs) have proved their effectiveness on a variety of domains and tasks. However, it has been found that DNNs are highly vulnerable to adversarial examples  \cite{szegedy2013intriguing}. To enhance the robustness of DNNs against adversarial examples, adversarial training \cite{goodfellow2014explaining,madry2017towards} has become one of the most effective and widely used methods. Given a pre-defined perturbation tolerance, denoted as $\epsilon$, adversarial training aims to minimize the robust loss, defined as the worst-case loss within $\epsilon$-ball around each example, leading to a min-max optimization problem. 
\cite{madry2017towards} show that applying a multi-step projected gradient descent (PGD) attack to approximately solve the inner maximization leads to a robust model, and several recent research has proposed various ways to improve adversarial training~\cite{zhang2019theoretically,wang2018bilateral,wang2019convergence,balaji2019instance,ding2018max}. 

%

However, the standard adversarial training methods still have a hypothetical and possibly problematic assumption: the perturbation tolerance $\epsilon$ is a large and fixed constant throughout the training process, which ignores the fact that  every data point may have different intrinsic robustness. Intuitively, some examples are naturally closer to the decision boundary, and enforcing large margin on those examples will force the classifier to give up on those examples, leading to a distorted decision surface.
This intuition may explain the known issue of the undesirable robustness-accuracy tradeoff in adversarial robustness \cite{su2018robustness,tsipras2019robustness}. 
Furthermore, with a different perturbation tolerance, it is questionable whether we should still force the model to learn to fit the one-hot label as in the original adversarial training formulation. In the extreme case, if an example is perturbed to the decision boundary, a good classifier yielding the binary class prediction probabilities should output $[0.5, 0.5]$ instead of $[1,0]$. This point becomes crucial when each example is associated with a different level of perturbation. 
Although some recent papers have started to address the uniform $\epsilon$ issue by treating correctly and incorrectly classified examples differently~\cite{ding2018max} or assigning non-uniform perturbation level~\cite{balaji2019instance}, none of them have tried to incorporate the customized training labels in this process. 
%


Motivated by these ideas, we propose a novel Customized Adversarial Training (CAT) framework that can substantially improve the performance of adversarial training. Throughout the adversarial training process, our algorithm dynamically finds a non-uniform and effective perturbation level and the  corresponding customized target label for each example. 
This leads to better generalization performance and furthermore, with a careful design on adaptive $\epsilon$ tuning, our algorithm has only negligible computational overhead and runs as fast as the original adversarial training algorithm.  Furthermore, we theoretically explain why the proposed method could lead to improved generalization performance. 

Our method significantly outperforms existing adversarial training methods on the standard CIFAR-10 defense task. With Wide-ResNet structure on CIFAR-10, under $8/255$ $\ell_\infty$ perturbation, our method achieves 73\% robust accuracy under PGD attack and 71\% robust accuracy under Carlini and Wagner (C\&W) attack \cite{carlini2017towards}, while the current best model only achieves 58.6\% under PGD attack and 56.8\% under C\&W attack. Furthermore, our method only degrades the clean accuracy from 95.93\% (standard test accuracy) to 93.48\%, while other adversarial training methods have clean accuracy below 91.34\%.

%
%


\section{Related Work}

\paragraph{Adversarial attack.}
Finding adversarial examples, also known as adversarial attacks, can be formulated as an optimization problem --- the goal is to find the perturbation $\delta$ to maximize the (robust) loss, while constraining $\delta$ to have small norm (e.g., $\ell_p$ norm). Therefore gradient-based algorithms have been widely used, such as fast gradient sign method (FGSM)~\cite{goodfellow2014explaining}, C\&W attack~\cite{carlini2017towards} and PGD attack~\cite{madry2017towards}. In addition to white-box attacks, it has been also found that adversarial attacks can be generated also in the soft-label black box setting~\cite{chen2017zoo,ilyas2018black} and hard-label black box setting~\cite{brendel2017decision,cheng2018query,cheng2020signopt}, and with similar quality to white-box attacks. Moreover, physical attacks have been proposed to generate adversarial examples in the real world~\cite{eykholt2018robust}.  Therefore, with the existence of these powerful adversarial attacks, how to enhance the robustness of neural network models has become an important issue in many real world applications. 

\paragraph{Adversarial training.}
To enhance the adversarial robustness of a neural network model, a natural idea is to iteratively generate adversarial examples, add them back to the training data, and retrain the model. 
For example, \citet{goodfellow2014explaining} use adversarial examples generated by FGSM to augment the data, and  \citet{kurakin2016adversarial_ICLR} propose to use a multiple-step FGSM to further improve the performance. 
\citet{madry2017towards} show that adversarial training can be formulated as a min-max optimization problem, and propose to use PGD attack (similar to multi-step FGSM) to find adversarial examples for each batch. The resulting method achieves notable successes and can survive even under strong attacks~\cite{athalye2018obfuscated}. 
After that, many defense algorithms are based on a similar min-max framework. 
\citet{zhang2019theoretically} propose TRADES, a theoretically-driven upper bound minimization algorithm to achieve the top-1 rank in NeurIPS 2018 defense competition. Recently, \citet{ding2018max} notice the importance of  misclassified examples and treat correctly classified and misclassified examples differently. \citet{zou2018improving} use label prediction probability as a smooth way to combine correctly and misclassified samples. Other than just adding adversarial examples into the training process, \citet{wang2018bilateral} find that it is also effective to find the ``adversarial label'' along with the ``adversarial perturbation''. The convergence of adversarial training has also been studied~\cite{gao2019convergence,wang2019convergence}.  Recently, to reduce the computational overhead brought by adversarial training, several works have been proposed \citep{shafahi2019adversarial,zhang2019you,wong2020fast}. It is widely recognized that the current defensive models are still not ideal and have considerable room for improvement. Moreover, to make robust models useful in practice, it is crucial that both clean and robust error need to be further enhanced. 

\paragraph{Other adversarial defenses}
In addition to adversarial training based methods, a wide range of defense methods have been proposed such as Gaussian data augmentation \cite{zantedeschi2017efficient}, randomized smoothing \cite{liu2018towards,cohen2019certified}, Mixup \cite{zhang2017mixup} and its variants \cite{thulasidasan2019mixup,verma2018manifold}, and Label smoothing \citep{shafahi2018label,goibert2019adversarial}. \citet{shafahi2018label} find that it could achieve similar robust accuracy with adversarial training when combining Gaussian data augmentation and label-smoothing.

However, some of the aforementioned methods have been shown to cause obfuscated gradients instead of enhanced robustness \cite{athalye2018obfuscated}, while adversarial training based methods are still shown to be robust under different kinds of adversarial attacks.

\section{Proposed Method}

\subsection{Preliminaries}
Adversarial training can be formulated as a min-max optimization problem. 
For a $K$-class classification problem, let $\cD = \{(\bx_i, y_i)\}_{i=1,\dots,n}$ denote the set of training samples in the dataset with $\bx_i\in\RR^d, y_i\in\{1,\dots,K\}=:[K]$. Let  $f_\theta(\bx): \RR^d  \rightarrow [K]$ denote a classification model parameterized by $\theta$. We denote by $h_\theta(\bx): \RR^d\rightarrow [0,1]^K$ as the prediction output for each class, i.e., $f_\theta(\bx)=\argmax_{i} [h_\theta(\bx)]_i$. We use standard $O(\cdot)$ notation to hide universal constant factor, and $a\lesssim b$ to indicate $a=O(b)$. 

Adversarial training can be formulated as:
    \begin{equation}
        \min_{\theta} \frac{1}{n} \sum_{i=1}^n \max_{\bx_i' \in \cB(\bx_i, \epsilon)} \ell(f_\theta(\bx_i'), y_i),
    \end{equation}
where $\cB(\bx_i,\epsilon)$ denotes the $\ell_p$-norm ball centered at $\bx_i$ with radius $\epsilon$. The inner maximization problem aims to find an adversarial version of a given data point $\bx_i$ that achieves a high loss. 
In general one can define $\cB(\bx_i,\epsilon)$ based on the threat model, but the $\ell_\infty$ ball is the most popular choice adopted by recent works \cite{madry2017towards,zhang2019theoretically,wang2018bilateral,ding2018max,zou2018improving}, which will also be used in this paper. 
For a deep neural network model, the inner maximization does not have a closed form solution, so adversarial training methods typically use a gradient-based iterative solver to approximately solve the inner problem. The most commonly used choice is the multi-step PGD~\cite{madry2017towards} and C\&W attack~\cite{carlini2017towards}. 

\begin{figure*}[t]
    \centering
    \begin{tabular}{cccc}
        \subfloat[Standard training\label{fig:linear_clean}]{\includegraphics[width=0.23\textwidth]{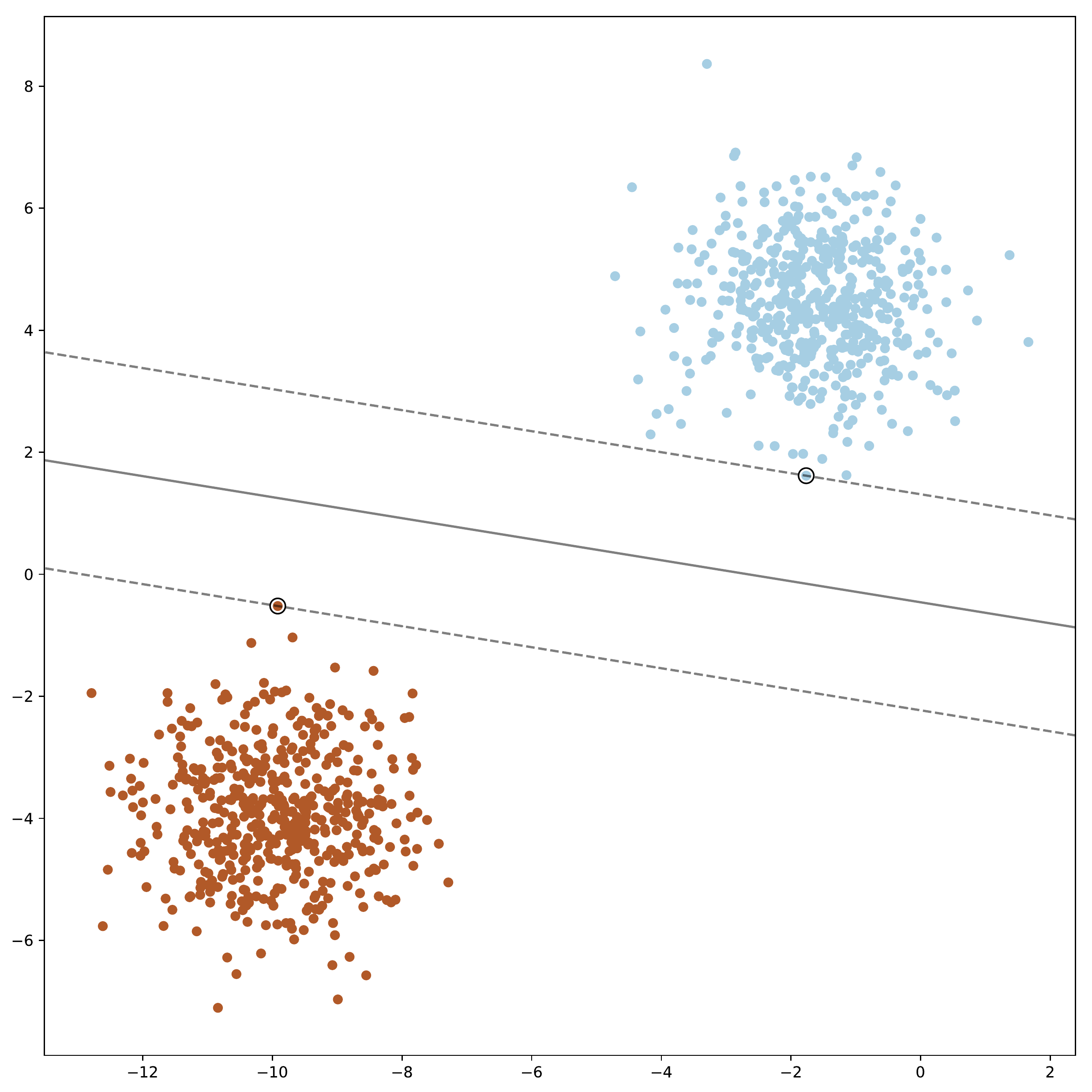}}
		&
        \subfloat[Adv-train with $\epsilon=1$\label{fig:linear_adv_small}]{\includegraphics[width=0.23\textwidth]{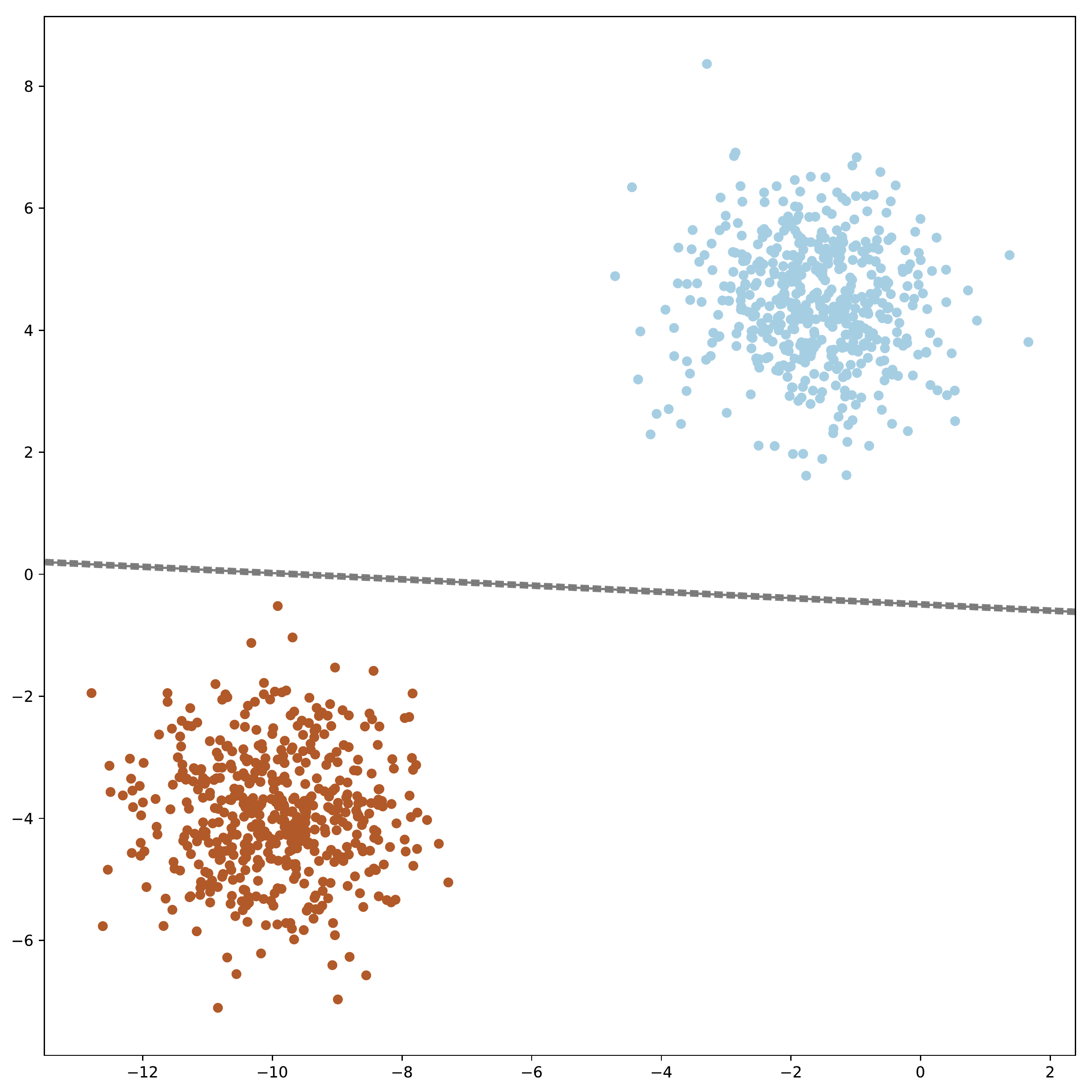}}
		&
		\subfloat[Adv-train with $\epsilon=4$\label{fig:linear_adv_large}]{\includegraphics[width=0.23\textwidth]{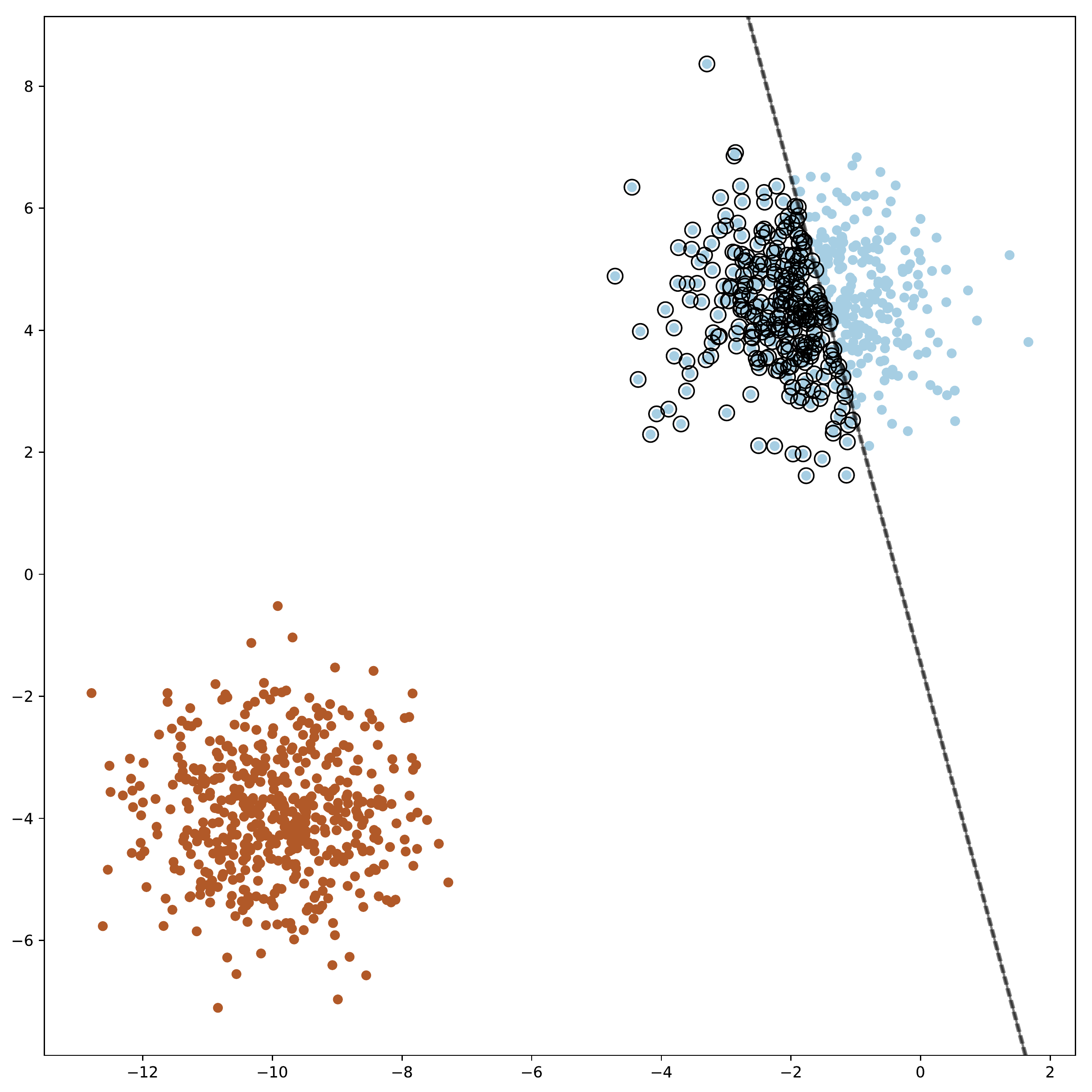}}
		&
		\subfloat[CAT (ours) with $\epsilon_{max}=4$ \label{fig:linear_cat}]{\includegraphics[width=0.23\textwidth]{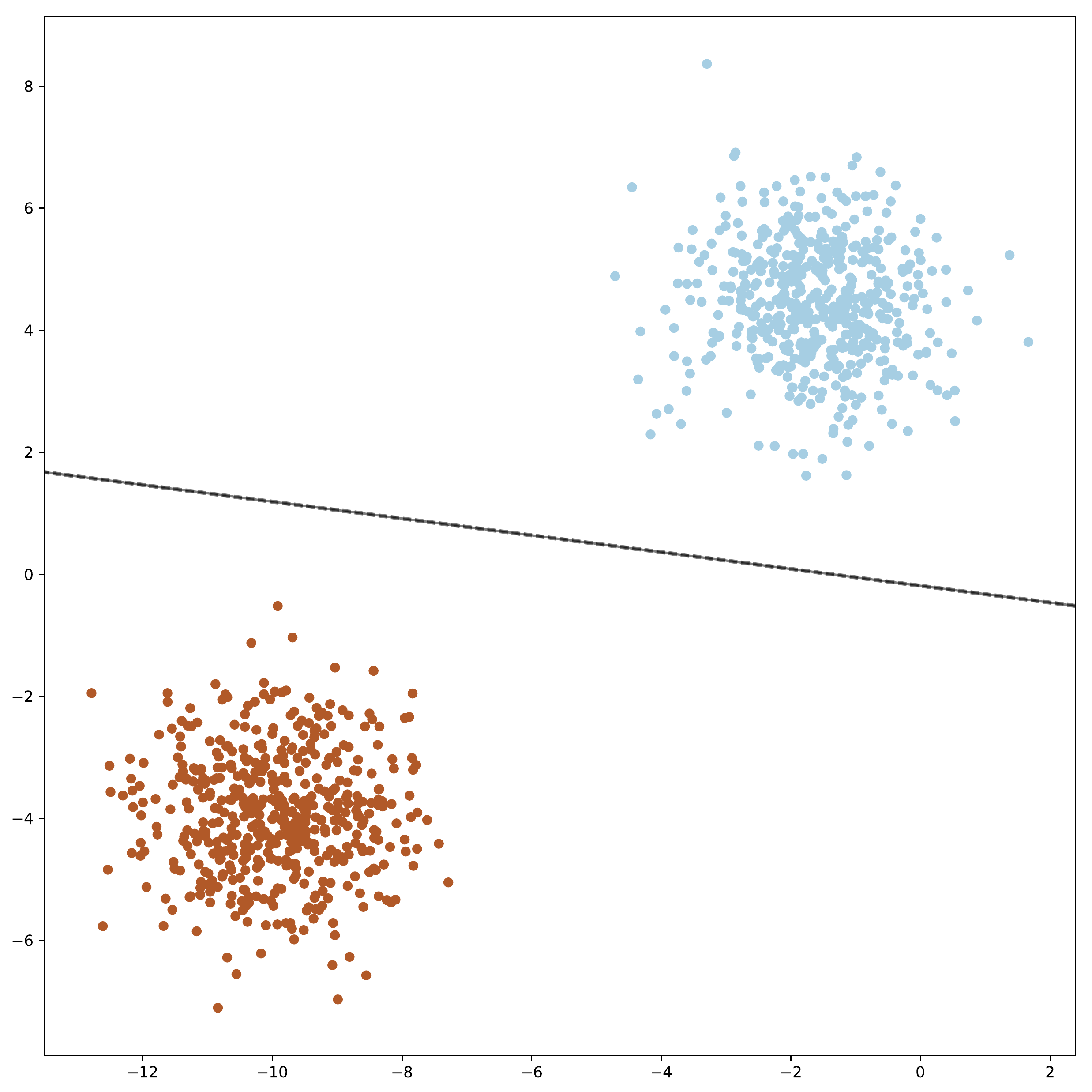}}
	\end{tabular}
		\caption{Different training methods on a linearly separable binary classification dataset with $1.75$ margin for both classes.
		Adversarial training with small $\epsilon$ works fine, but for a large $\epsilon$ beyond the true margin, 
		adversarial training would ruin the classifier's classification performance, while our proposed adaptive customized adversarial training method still keeps a good generalization performance. 
		}
		\label{fig:linear}
\end{figure*}

\subsection{Motivation}
\label{sec:motivation}
Intuitively, if adversarial training can always find a model with close-to-zero robust error, one should always use a large $\epsilon$ for training because it will automatically imply robustness to any smaller $\epsilon$. Unfortunately, in practice a uniformly large $\epsilon$ is often harmful. In the following we empirically explain this problem  and use it to motivate our proposed algorithm. 

We use a simple linear classification case to demonstrate why a uniformly large $\epsilon$ is harmful. In Figure \ref{fig:linear_clean}, we generate a synthetic linearly separable dataset with the margin set to be 1.75 for both classes, and the correct linear boundary can be easily obtained by standard training. In Figure \ref{fig:linear_adv_small}, we run adversarial training with $\epsilon=1$, and since this $\epsilon$ is smaller than the margin, the algorithm can still obtain near-optimal results. However, when we use a large $\epsilon=4$ for adversarial training in Figure \ref{fig:linear_adv_large}, the resulting decision boundary becomes significantly worse. It is because adversarial training cannot correctly fit all the samples with a margin up to 4, so it will sacrifice some data samples, leading to distorted and undesirable decision boundary. This motivates the following two problems:
\begin{compactitem}
    \item We shouldn't set the same large $\epsilon$ uniformly for all samples. Some samples are intrinsically closer to the decision boundary and they should use a smaller $\epsilon$. Without doing this, adversarial training will give up on those samples, which leads to worse training and generalization error (see more discussions in Section \ref{sec:theory} on the generalization bounds). 
    \item The adversarial training loss is trying to force the prediction to match the one-hot label (e.g., $[1,0]$ in the binary classification case) even after large perturbations. However, if a sample is perturbed, the prediction shouldn't remain one-hot. For instance, if a sample is perturbed to the decision boundary, the prediction of a perfect model should be $[0.5, 0.5]$ instead of $[1, 0]$. This also makes adversarial training fail to recover a good decision hyperplane. 
\end{compactitem}
Furthermore, we observe that even if adversarial training can obtain close-to-zero training error with large $\epsilon$ (e.g., \citep{gao2019convergence} proves that this will happen for overparameterized network with large-enough margin), a uniformly large $\epsilon$ will lead to larger generalization gap. 
This could be partially explained by the theoretical results provided by~\cite{yin2018rademacher}, which shows that the adversarial Rademacher complexity has a lower bound with an explicit dependence on the perturbation tolerance. 
The empirical results in Table~\ref{tab:motivation} also illustrate this problem. When conducting adversarial training with $\epsilon=0.3$ on CIFAR10 VGG-16, we found that the model achieves close-to-zero robust training error on all $\epsilon\leq 0.3$, but it suffers larger generalization gap compared to training with smaller $\epsilon$. This also demonstrates that a uniformly large $\epsilon$ is harmful even when it achieves perfect training error.

\begin{table}[htbp]    
    \caption{The influence of different fixed $\epsilon$ values used in adversarial training on the robust accuracy with $\epsilon=0.01$. }
    \label{tab:motivation}
    \centering
    \begin{tabular}{|c|c|c|c|c|}
    \hline
         \multirow{2}{*}{Testing $\epsilon$} &\multirow{2}{*}{Error Type} & \multicolumn{3}{c|}{Training $\epsilon$}  \\ \cline{3-5} 
           & & 0.01 & 0.02 & 0.03   \\
           \hline
            \multirow{2}{*}{0.01} & Train& 99.96\% & 99.99\% & 99.16\%\\ \cline{2-5}
            & Test& 69.79\% & 69.06\% & 66.04\%\\
           \hline
    \end{tabular}
\end{table}

\paragraph{CAT (Customized Adversarial Training)} 
We propose the Customized Adversarial Training (CAT) framework that improves adversarial training by addressing the above-mentioned problems. First, our algorithm has an auto-tuning $\epsilon$ method to customize the $\epsilon$ used for each training example. Second, instead of forcing the model to fit the original label, we customize the target label for each example based on its own $\epsilon$. In the following we will describe these two components in more detail.




\subsection{Auto-tuning $\epsilon$ for adversarial training}
\label{sec:epsilon}

The first component of our algorithm is an auto-tuning $\epsilon$ tuning method which adaptively assigns a suitable $\epsilon$ for each sample during the adversarial training procedure. 
Let $\epsilon_i$ be the perturbation level assigned to example $i$. 
Based on the intuition mentioned in Section \ref{sec:motivation}, we do not want to further increase $\epsilon$ if we find the classifier does not have capacity to robustly classify the example, which means we should set
\begin{align}
 \epsilon_i = \argmin_{\epsilon} \{\max_{\bx_i' \in \cB(\bx_i, \epsilon)}f_\theta(\bx_i')\neq y_i\}
 \label{eq:epsilon}
\end{align}
and the adversarial training objective becomes 
\begin{align}
\min_{\theta}  \frac{1}{n} \sum_{i=1}^n \max_{\bx_i' \in \cB(\bx_i, \epsilon_i)} \ell(f_\theta(\bx_i'), y_i).
\label{eq:adaptive}
\end{align}
Note that $\epsilon_i$ in \eqref{eq:epsilon} depends on $\theta$ while $\theta$ in \eqref{eq:adaptive} also depends on $\epsilon_i$. 
We thus propose to conduct alternative updates --- conducting one SGD update on $\theta$, and then update the $\epsilon_i$ in the current batch. However, finding $\epsilon_i$ exactly requires brute-force search for every possible value, which adds significant computational overhead to adversarial training. 
Therefore, we only conduct a simplified update rule on $\epsilon_i$ as follows. Starting from an initial perturbation level of zero, at each iteration
we conduct adversarial attack (e.g., PGD attack) with perturbation tolerance $\epsilon_i + \eta$ where $\eta $ is a constant. If the attack is successful, then we keep the current $\epsilon_i$, while if the attack is unsuccessful, which means an attacker still cannot find an adversarial example that satisfies $\max_{\bx_i' \in \cB(\bx_i, \epsilon_i+\eta)}f_\theta(\bx_i')\neq y_i$, then we increase $\epsilon_i=\epsilon_i+\eta$. The attack results will also be used to update the model parameter $\theta$, so this adaptive scheme does not require any additional cost. 
In practice, we also have an upper bound on the final perturbation to make sure each individual $\epsilon_i$ will not be too large. 

%

\subsection{Adaptive label uncertainty for adversarial training}
\label{sec:label}


As mentioned in Section~\ref{sec:motivation}, the standard adversarial training loss is trying to enforce a sample being classified as the original one-hot label after $\epsilon$ perturbation. However, this may not be ideal. In the extreme case, if a sample is perturbed to the decision boundary, the prediction must be far away from one-hot. 
This problem is more severe when using non-uniform $\epsilon_i$, since each different $\epsilon_i$  will introduce a different bias to the loss, and that may be one of the reasons that purely adaptive $\epsilon$-scheduling does not work well (see our ablation study in Section~\ref{sec:aba} and also the results reported in~\cite{balaji2019instance}).


In the following, we propose an adaptive label smoothing approach to reflect different perturbation tolerance on each example. 
\citet{szegedy2016rethinking} introduced label smoothing that converts one-hot label vectors into one-warm vectors representing low-confidence classification, in order to prevent the model from making over-confident predictions. Specifically, with a one-hot encoded label $y$, the smoothed version is 
\begin{equation*}
    \tilde{y} = (1-\alpha)y+\alpha u
    \end{equation*}
where $\alpha \in [0,1]$ is the hyperparameter to control the smoothing level. In the adaptive setting, we set $\alpha=c*\epsilon_i$ so that a larger perturbation tolerance would receive a higher label uncertainty and $c$ is a hyperparameter. A common choice of $u$ is the uniform distribution $u=\frac{1}{K}$. To further prevent over-fitting and improve the generalization, we use $u=\text{Dirichlet}(\one)$ where Dirichlet($\cdot$) refers to the Dirichlet distribution and $\one \in \RR^K$ is an all one vector. With different perturbation tolerance, the adaptive version of label smoothing is 
\begin{equation}
    \tilde{y_i} = (1-c\epsilon_i)y_i+c\epsilon_i\text{Dirichlet}(\one).
    \label{eq:tildey_new}
\end{equation}    

\textbf{The final objective function}:
Combining the two aforementioned two main techniques, our Customized Adversarial Training (CAT) method attempts to minimize the following objective: 
\begin{align}
\begin{split}
\min_{\theta} & \frac{1}{n} \sum_{i=1}^n \max_{\bx_i' \in \cB(\bx_i, \epsilon_i)} \ell(f_\theta(\bx_i'), \tilde{y_i})\\
& \textbf{s.t.}\; \epsilon_i = \argmin_{\epsilon} \{\max_{\bx_i' \in \cB(\bx_i, \epsilon)}f_\theta(\bx_i')\neq y_i\}
\end{split}
\end{align}
where $\tilde{y_i}$ is defined in \eqref{eq:tildey_new}.
As described in Section \ref{sec:epsilon}, we approximately minimize this objective with an alternative update scheme, which encounters  almost no additional cost compared to the original adversarial training algorithm. The detailed algorithm is shown in Algorithm \ref{alg:cat}.

\textbf{Choice of loss function. }
In general, our framework can be used with any loss function $\ell(\cdot)$. 
In the previous works, cross entropy loss is commonly used for $\ell$. However, the model trained by smoothing techniques tends to have better performance against PGD attack than C$\&$W$_\infty$ attack (see the VGG experiments in Figure~\ref{fig:res}.
 So in addition to testing our algorithm under cross entropy loss, we also propose a mixed loss to enhance the defense performance towards C$\&$W$_\infty$ attack. That is,
\begin{equation}
    \text{CE}(x_i, \tilde{y_i}) + \max\{[\max_{i\neq y_0}[Z(\bx)]_i-[Z(\bx)]_{y_0},-\kappa\},
    \label{eq:mix}
\end{equation}
where $Z(\bx) \in \RR^K$ is the final (logit) layer output, and $[Z(\bx)]_i$
is the prediction score for the i-th class and $y_0$ is the original label. The parameter $\kappa$ encourages to find an adversary that will not classified as class $y_0$ with high confidence.

\begin{algorithm}
   \caption{CAT algorithm}
   \label{alg:cat}
\begin{algorithmic}
   \STATE {\bfseries Input:} Training dataset $(X,Y)$, cross entropy loss or mix loss $\ell$, scheduling parameter $\eta$, weighting factor $c$,
   perturbation upperbound  $\epsilon_{max}$
   \STATE Initial every sample's $\epsilon_i$ with 0

   \FOR{epoch=$1,\dots,N$}
   \FOR{i=$1,\dots,B$}
   \STATE $\tilde{y_i} \leftarrow (1-c\epsilon_i) y_i+(1-c\epsilon_i)\text{Dirichlet}(\one)$
    \STATE $\epsilon_i \leftarrow \epsilon_i + \eta$
    \STATE $\delta_i \leftarrow 0$
   \FOR{$j=1\dots m$}
   \STATE $\delta_i \leftarrow \delta_i + \alpha \cdot sign(\nabla_\delta\ell(f_\theta(\bx_i+\delta_i),\tilde{y_i})$
   \STATE $\delta_i \leftarrow \max(\min(\delta_i,\epsilon_i),-\epsilon_i)$
    \ENDFOR
    \IF{$f_\theta(\bx_i+\delta_i)\neq y_i$}
    \STATE $\epsilon_i \leftarrow \epsilon_i - \eta$
    \ENDIF
   \STATE $\epsilon_i \leftarrow \min(\epsilon_{max},\epsilon_i)$
   \STATE $\tilde{y_i} \leftarrow (1-c\epsilon_i) y_i+(1-c\epsilon_i)\text{Dirichlet}(\one)$
   \STATE $\theta \leftarrow \theta - \gamma_\theta \nabla_\theta \ell(f_\theta(\bx_i+\delta_i),\tilde{y_i})$ 
   \ENDFOR
   \ENDFOR
   \STATE {\bfseries return} $\theta$
\end{algorithmic}
\end{algorithm}

\subsection{Theoretical Analysis}
\label{sec:theory}
To better understand how our scheme improves generalization, we provide some theoretical analysis. Recall we denote by  $h_\theta(\bx): \RR^d \rightarrow [0,1]^K$ as the prediction probability for the $K$ classes. We define the bilateral margin that our paper is essentially maximizing over as follows. 
\begin{definition}[Bilateral margin]
We define the bilateral perturbed network output by $H_\theta(\bx,\bdelta^{i},\bdelta^o)$:
\begin{align*}
H_\theta(\bx,\bdelta^i,\bdelta^o):=h_\theta\left(\bx+\bdelta^i\|\bx\|\right)+\bdelta^o\left\|\bx+\bdelta^{i}\|\bx\|\right\|.     
\end{align*}
The bilateral margin is now defined as the minimum norm of $(\bdelta^i,\bdelta^o)$ required to cause the classifier to make false predictions: 
\begin{equation}
    \begin{aligned}
    \label{eqn:margin}
    m_F(\bx,y):=& \min_{\bdelta^i, \bdelta^o} \sqrt{ \|\bdelta^i\|^2+\|\bdelta^o\|^2}\\
    &\text{ subject to } \max_{y'} H_\theta(\bx,\bdelta^i,\bdelta^o)_{y'}\neq y. 
    \end{aligned} 
\end{equation}
\end{definition} 
This margin captures both the relative perturbation on the input layer $\bdelta^i$ and on the soft-max output $\bdelta^o$. 
\begin{theorem}
\label{thm:main} 
Suppose the parameter space $\Theta$ we optimize over  has covering number that scales as $\log \cN_{\|\cdot\|_{\text{op}}}(\eta,\Theta)\leq \lfloor \cC^2/\eta^2\rfloor$ for some complexity $\cC$. Then with probability $1-\delta$ over the draw of the training data, any classifer $f_{\theta}, \theta\in \Theta$ which achieves training error 0 satisfies:
\begin{equation*}
    \mathbb{E}[f_\theta(\bx)=y] \lesssim \frac{\cC \log^2 n}{\sqrt{n}}\sqrt{\frac{1}{n}\sum_{i=1}^n \frac{1}{m_F(\bx_i,y_i)} } + \zeta,
\end{equation*}
where $\zeta$ is of small order $O\left(\frac{1}{n}\log(1/\delta)\right)$.
\end{theorem}
We defer the proof to the Appendix, which is adapted from Theorem 2.1 of \cite{wei2019improved}. We observe the population risk is bounded by two key factors, the average of $ \frac{1}{m_F(\bx_i,y_i)}$ and $\cC$, the covering number of the parameter space. On one side, the average of $\frac{1}{m_F(\bx_i,y_i)}$ is dominated by the samples with the smallest margin. Therefore when we do adversarial training, it is important that we not only achieve higher overall accuracy, but also make sure the samples closer to the decision boundary have large enough margin. This can not be achieved by simply using constant and large $\epsilon$ that will maintain a large margin for most samples but sacrifice the accuracy of a small portion of data. On the other hand, the covering number of the network's parameter space can be roughly captured by a bound of product of all layers' weight norms. We hypothesize that with more flexibility in choosing $\epsilon$, our algorithm will converge faster than using larger constant $\epsilon$ and will have more implicit regularization effect. To testify this hypothesis, we roughly measure the model complexity $\cC$ by the product of the weight norms of different models. In comparison to our model, when training with constant $\epsilon = 0.01,0.02$ and $0.03$, it respectively yields $\cC$ as large as $2.54, 3.53$ and $ 1.39$ times of that of our model, which means our model indeed has more implicit regularization effect among others.



\subsection{Connections with other training methods}
Many recent papers attempt to improve adversarial training. Although they all follow the similar min-max framework, each of them uses slightly different loss functions. We summarize the loss functions used by recent adversarial training methods in Table \ref{tb:ref}. 


We see that except for natural training which directly minimizes  the cross entropy loss (denoted as CE), all training techniques involve the  use of the min-max framework. TRADES and MMA use the unperturbed data's cross entropy loss as an additional regularization term to achieve a better trade-off between clean and robust error. 

Similar to our method,  both MMA and IAAT have sample-wise adaptive $\epsilon$ during training. They also utilize the adaptive $\epsilon$ to find the largest possible $\epsilon_i$ for every sample $\bx_i$. However, they do not consider the adaptive label technique mentioned in Section \ref{sec:label}. As a result, they can only achieve better clean accuracy while the improvements in robust accuracy is limited. 
Our CAT-CE algorithm (CAT with CE loss) is more general than IAAT and MMA. CAT-CE reduce to IAAT when we set $c=0$ in adaptive label smoothing. Moreover, MMA could be treated as a special case of CAT-CE when we use a line search scheme to find the $\epsilon_i$ and $c=0$.
Also, in Section \ref{sec:aba}, we will show the importance of the adaptive label uncertainty step in CAT.

\begin{table*}[htbp]
\caption{Summary of several robust training methods amd their corresponding loss function. $\text{Dirichlet}(\mathbf{b})$ indicates the Dirichlet distribution parameterized by $\mathbf{b}$. 
}
\label{tb:ref}
\begin{center}
 \resizebox{.98\textwidth}{!}{
\begin{tabular}{lc}
\toprule
Methods & Loss Function \\
\midrule
Natural & $\text{CE}(f_\theta(\bx),y)$ \\ 
Adversarial training \citep{madry2017towards} & $\max_{\bx'\in \cB(\bx,\epsilon)}\text{CE}(f_\theta(\bx'),y)$ \\
TRADES \citep{zhang2019theoretically} & $\text{CE}(f_\theta(\bx),y)+ \max_{\bx'\in \cB(\bx,\epsilon)}\text{KL}(f_\theta(\bx'),f_\theta(\bx))$ \\
Bilateral Adv Training \citep{wang2018bilateral} & $\max_{\bx'\in \cB(\bx,\epsilon), y'\in \Delta}\text{CE}(f_\theta(\bx'),y')$ \\
MMA \citep{ding2018max} & $\text{CE}(f_\theta(\bx))\one(f_\theta(\bx)\neq y)+(\max_{\bx'\in \cB(\bx, \epsilon)}\text{CE}(f_\theta(\bx'),y))\one(f_\theta(\bx)=y)$ \\
MART \citep{zou2018improving} &  $\max_{\bx'\in \cB(\bx,\epsilon)}\text{BCE}(f_\theta(\bx'),y)+ \text{KL}(f_\theta(\bx'),f_\theta(\bx))\cdot (1-f_\theta(\bx))$\\
IAAT \citep{balaji2019instance} & $\max_{\bx_i' \in \cB(\bx_i, \epsilon_i)}\text{CE}(f_\theta(\bx_i'),y_i)$\\
CAT-CE (ours) & $\max_{\bx_i' \in \cB(\bx_i, \epsilon_i)}\text{CE}(f_\theta(\bx_i'),(1-c\epsilon_i)y_i+c\epsilon_i\text{Dirichlet}(\one))$ \\
CAT-MIX (ours) & $\max_{\bx_i' \in \cB(\bx_i, \epsilon_i)} \text{CE}(\bx'_i, (1-c\epsilon_i)y_i+c\epsilon_i\text{Dirichlet}(\one)) + \max_{j\neq y_0}[Z(\bx_i')]_j-[Z(\bx_i')]_{y_0}$ \\
\bottomrule
\end{tabular}
}
\end{center}
\end{table*}




\section{Performance Evaluation}
In this section, we conduct extensive experiments to show that CAT achieves a strong result on both clean and robust accuracy. We include the following methods into our comparison:
\begin{compactitem}
    \item Customized Adversarial Training (CAT-CE): Our proposed method with the cross entropy loss.
    \item Customized Adversarial Training (CAT-MIX): Our proposed method with the mixed cross entropy loss~\eqref{eq:mix}.
    \item Adversarial training: The adversarial training method proposed in \cite{madry2017towards} where they use a K-step PGD attack as adversary.
    \item TRADES: TRADES \citep{zhang2019theoretically} improves adversarial training by an additional loss on the clean examples and achieves the state-of-art performance on robust accuracy.
    \item Natural: the natural training which only minimizes the cross entropy loss.
\end{compactitem}
Furthermore, since  many recently proposed adversarial training  methods have considered  CIFAR-10 with Wide-ResNet structure as the standard setting and report their numbers, we also compare our performance with 7 previous methods on this specific setting.

\begin{figure*}[htbp]
    \centering
    \begin{tabular}{cc}
        \subfloat{\includegraphics[width=0.38\textwidth]{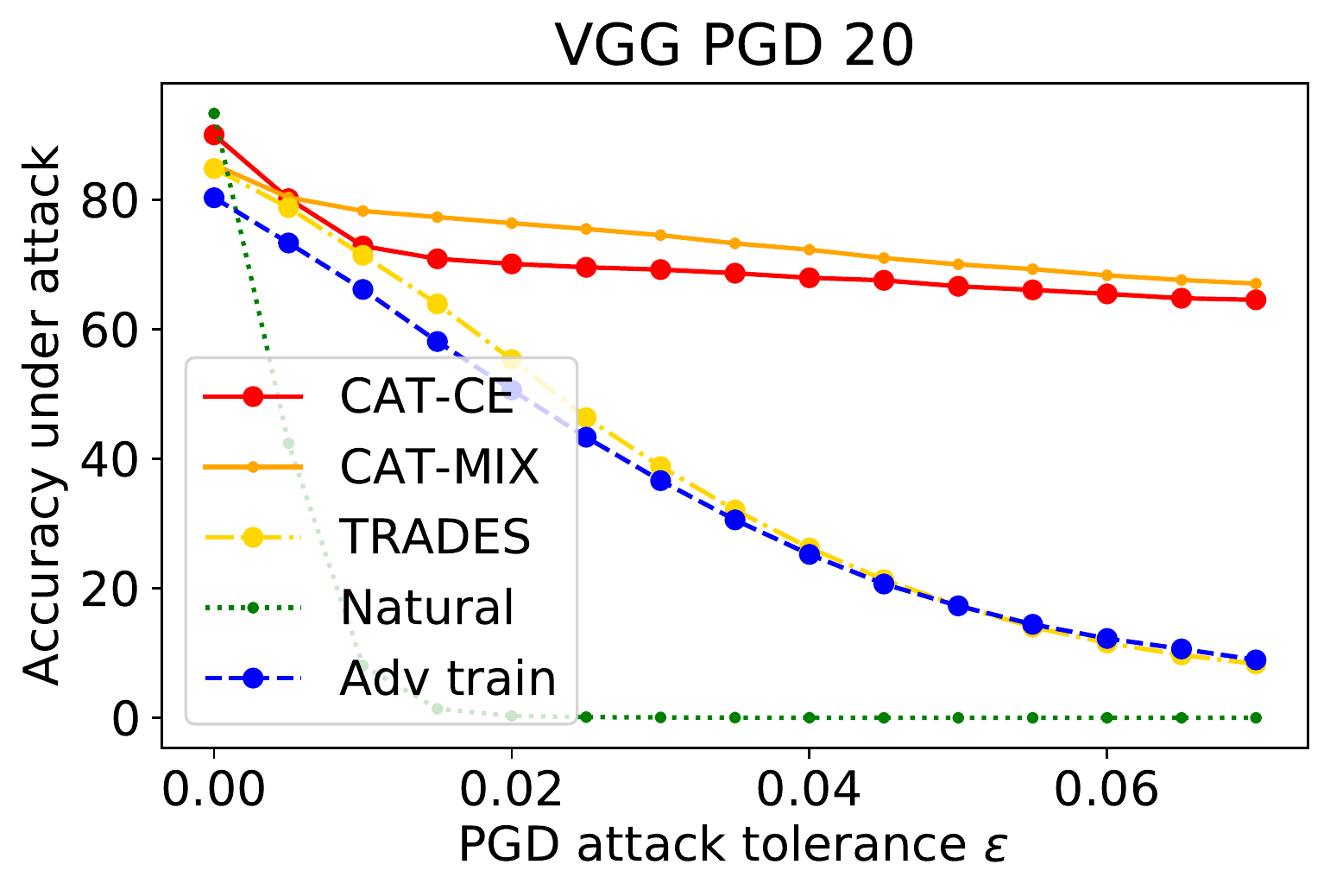}}
		&
		\subfloat{\includegraphics[width=0.38\textwidth]{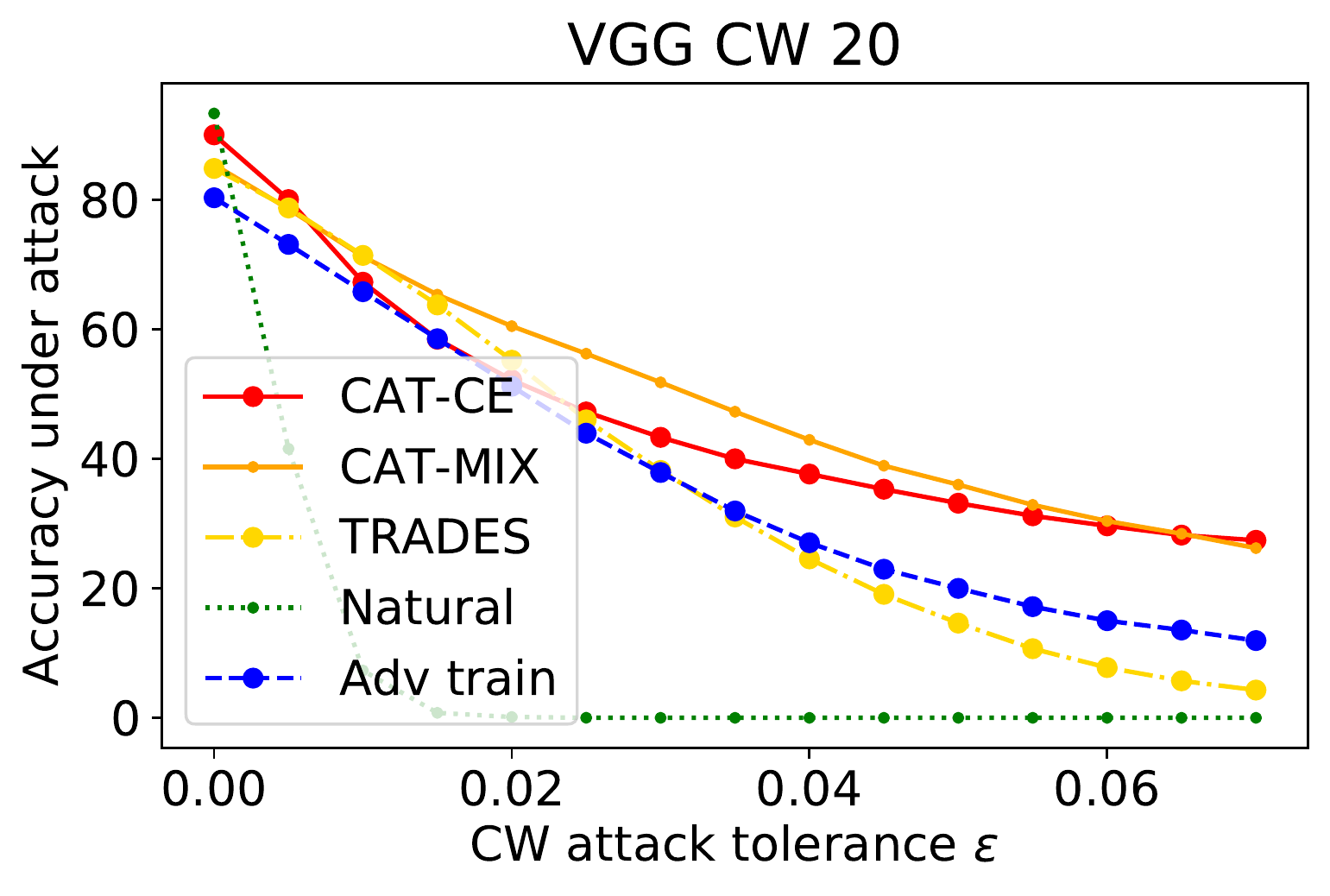}}
		\\
    \subfloat{\includegraphics[width=0.38\textwidth]{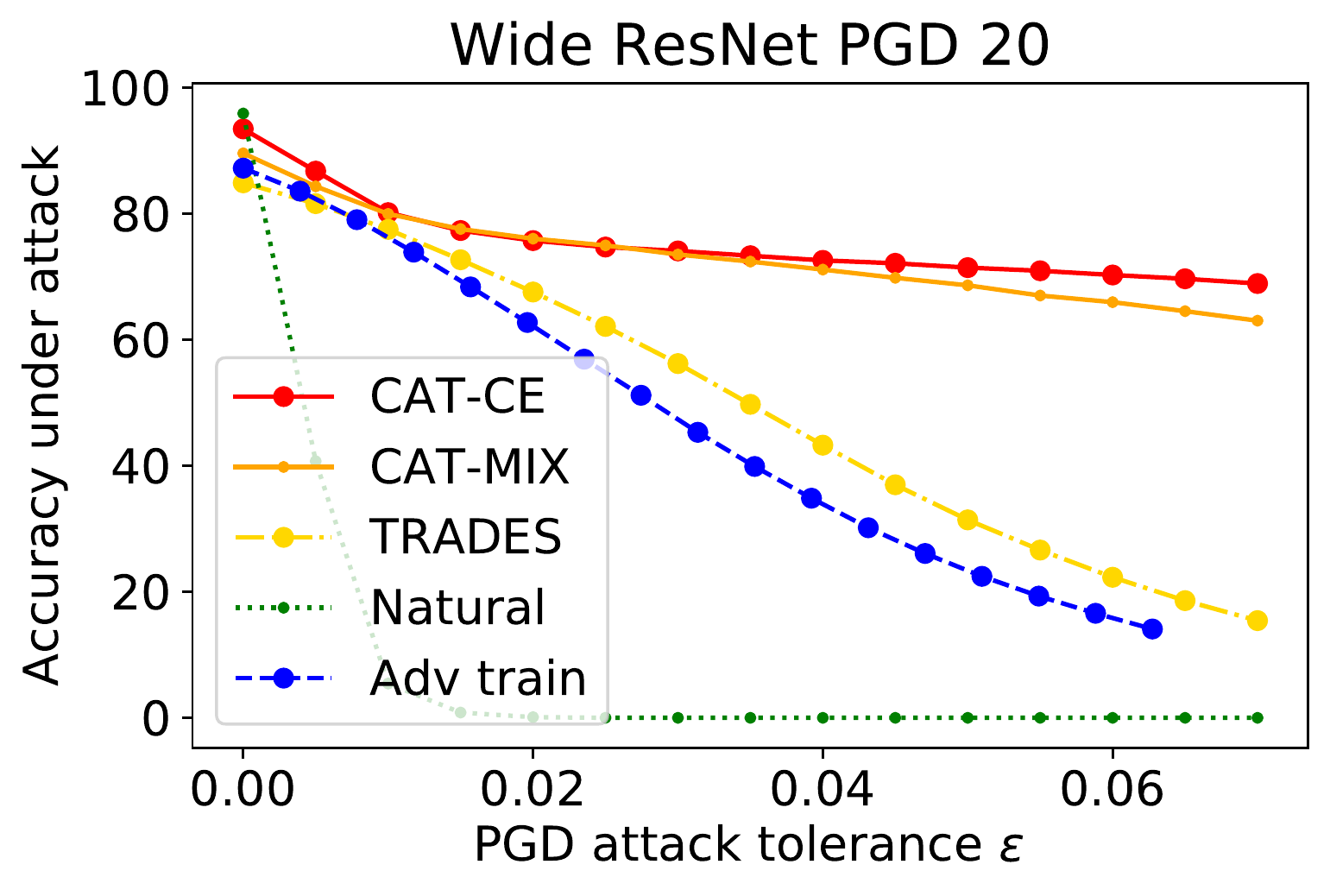}}
		&
		\subfloat{\includegraphics[width=0.38\textwidth]{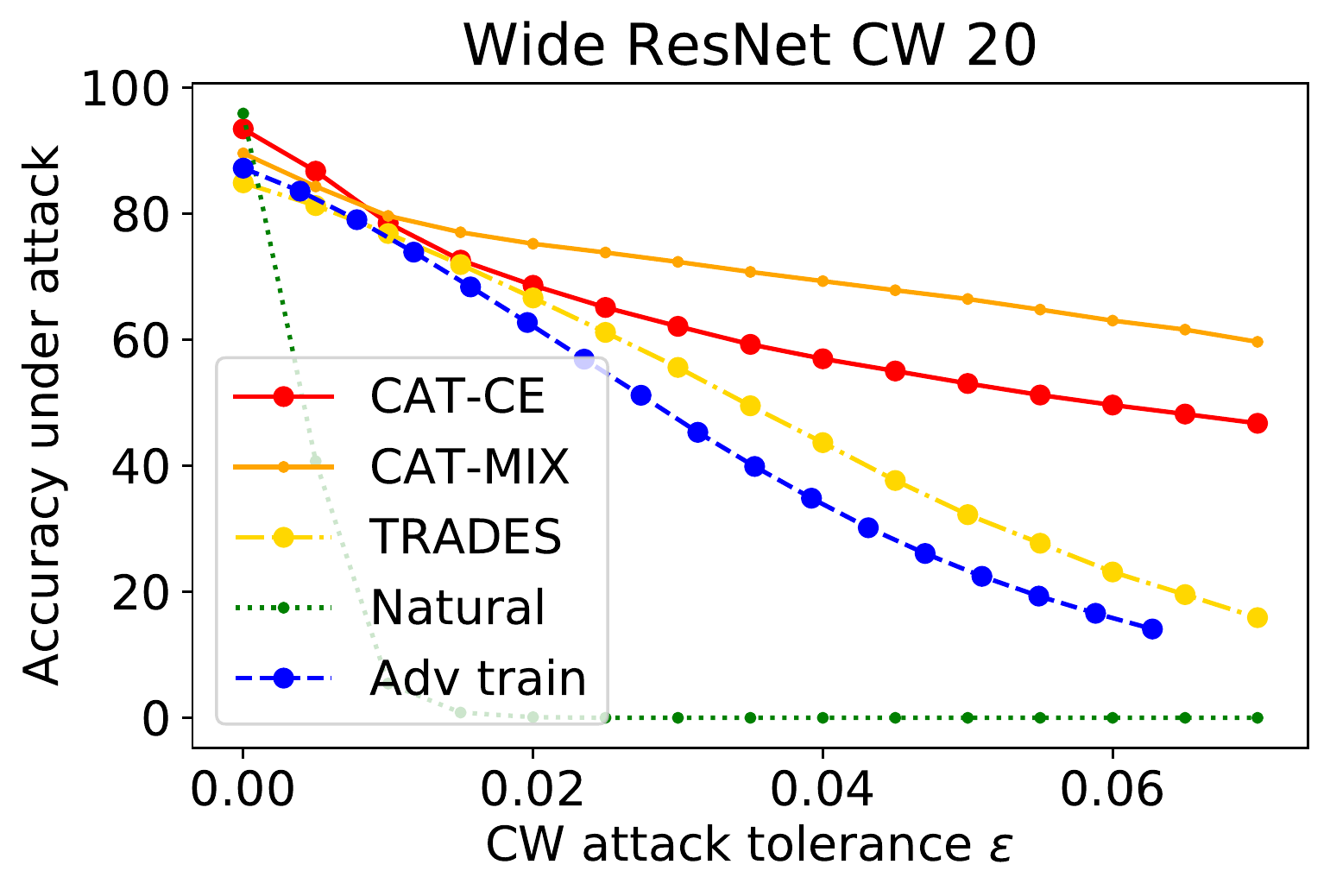}}
	\end{tabular}
		\caption{Robust accuracy under different levels of attacks on CIFAR-10 dataset with VGG and Wide-ResNet architectures. CAT-CE and CAT-MIX clearly outperform TREADS and adversarial training. 
		}
		\label{fig:res}
\end{figure*}

\begin{table*}[htbp]
\caption{The clean and robust accuracy of Wide Resnet models trained by various defense methods. All robust accuracy results use $\epsilon=8/255$  $\ell_\infty$  ball. We reported the best performance listed in the papers.
$^{(*)}$ denotes random-restart is applied in the testing attack. $^{(X)}$ denotes it use a $X$-step PGD attack }
\label{tb:main}
\begin{center}
\begin{tabular}{l|c|c|c}
\toprule
Methods & Clean accuracy & PGD accuracy & C$\&$W accuracy \\
\hline
Natural training & {\bf 95.93\%} & 0\% & 0\%\\ 
Adversarial training \citep{madry2017towards} & 87.30\% & 52.68\% & 50.73\% \\
Dynamic adversarial training \citep{wang2019convergence} & 84.51\% & 55.03\%& 51.98\%\\
TRADES \citep{zhang2019theoretically} &84.22\%& 56.40\%$^{(20)}$ & 51.98\% \\
Bilateral Adv Training \citep{wang2018bilateral} &91.00\% &57.5\%$^{(*20)}$  & 56.2\%$^{(*20)}$\\
MMA \citep{ding2018max} &84.36\% & 47.18\% & \xmark\\
MART \citep{zou2018improving} & 84.17\%& 58.56\%$^{(20)}$& 54.58\%\\
IAAT \citep{balaji2019instance} & 91.34\%& 48.53\%$^{(* 10)}$ & 56.80\% \\
CAT-CE (ours) & 93.48\% & {\bf 73.38\%}$^{(*20)}$ & 61.88\%$^{(*20)}$ \\
CAT-MIX (ours) & 89.61\% & 73.16\%$^{(*20)}$& {\bf 71.67\%}$^{(*20)}$\\
\bottomrule
\end{tabular}
\end{center}
\end{table*}
\begin{figure*}[htbp]
    \centering
    \begin{tabular}{cccc}
        \subfloat[Natural]{\includegraphics[width=0.225\textwidth]{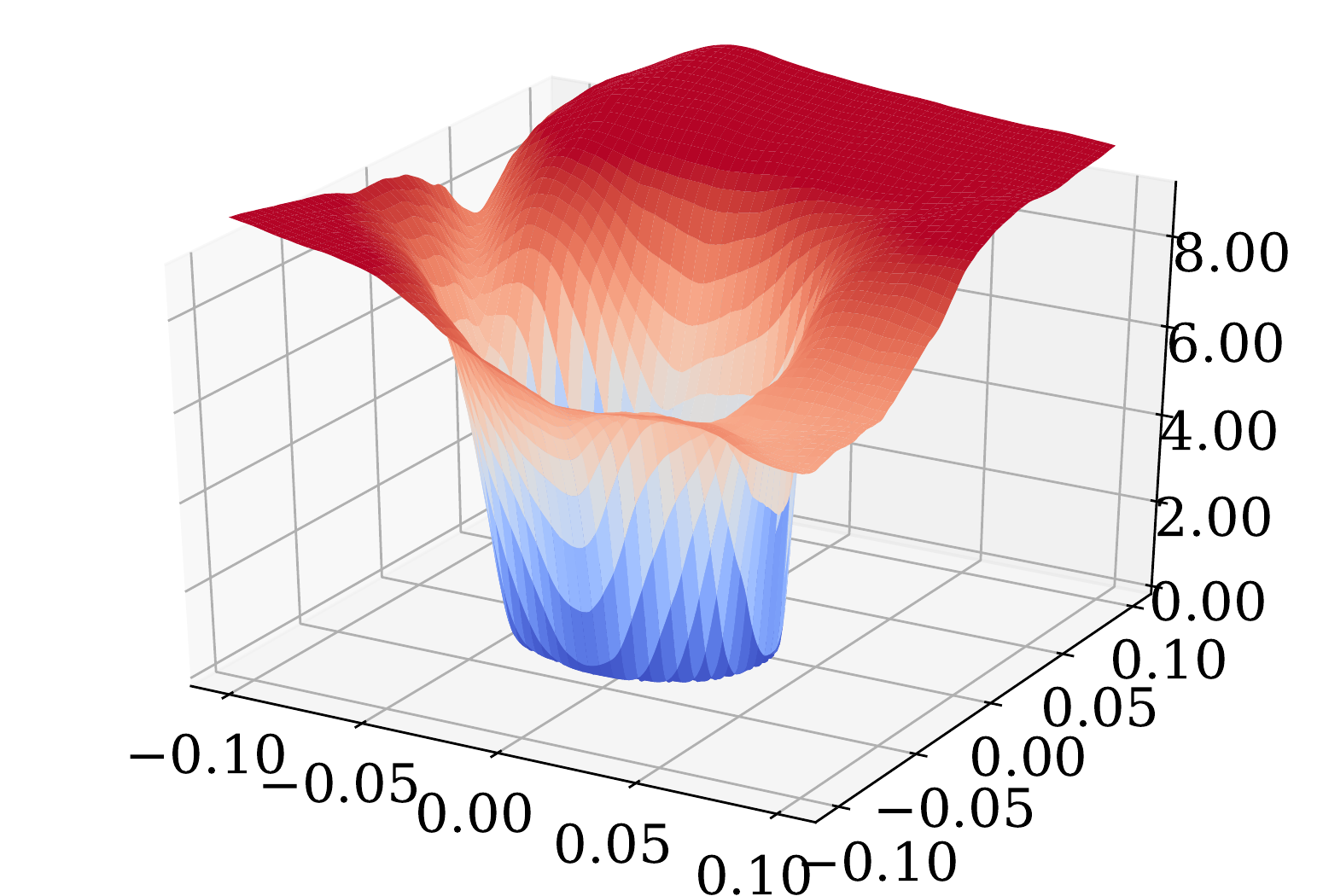}}
		&
		\subfloat[Adv train]{\includegraphics[width=0.225\textwidth]{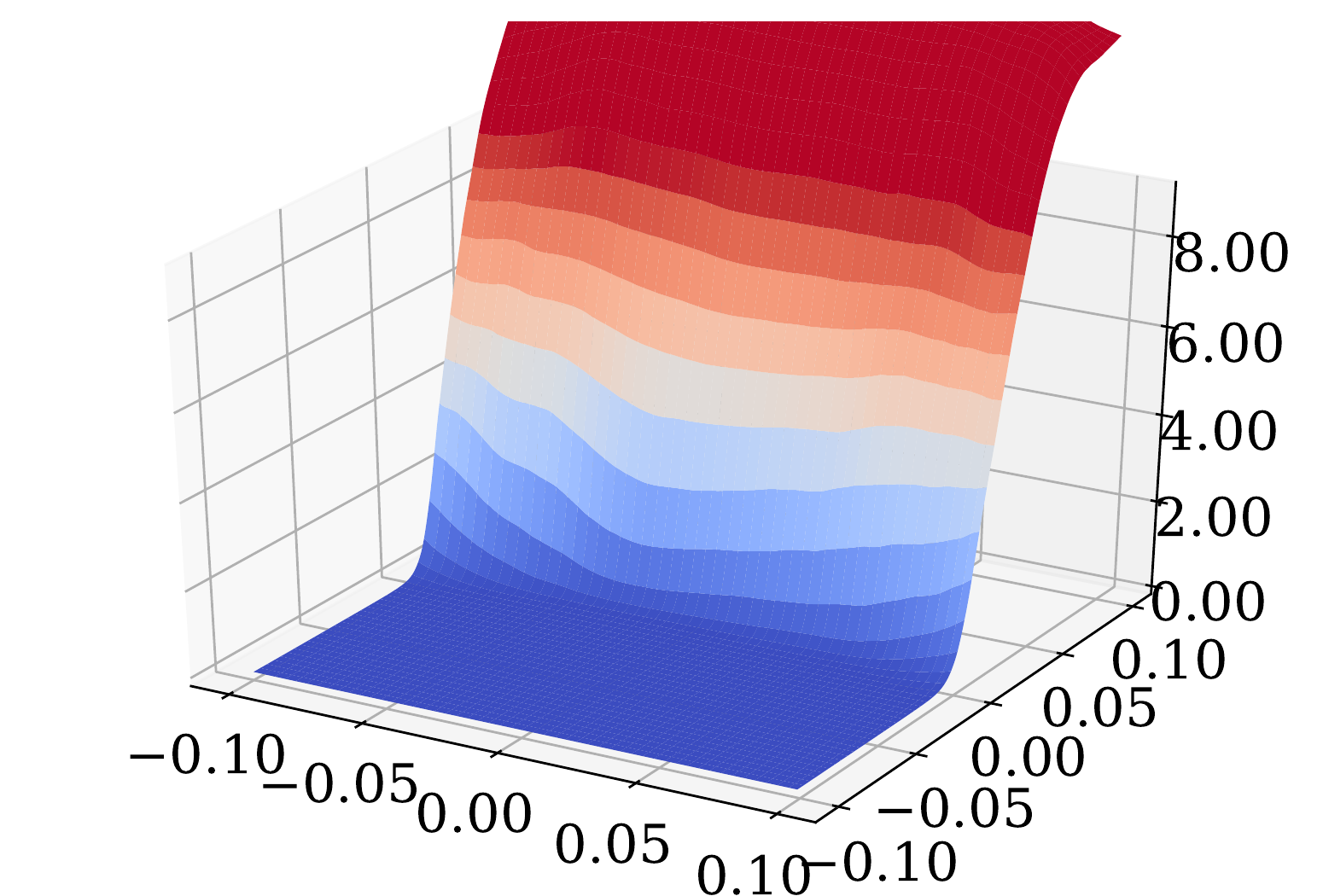}}
		&
        \subfloat[TRADES]{\includegraphics[width=0.225\textwidth]{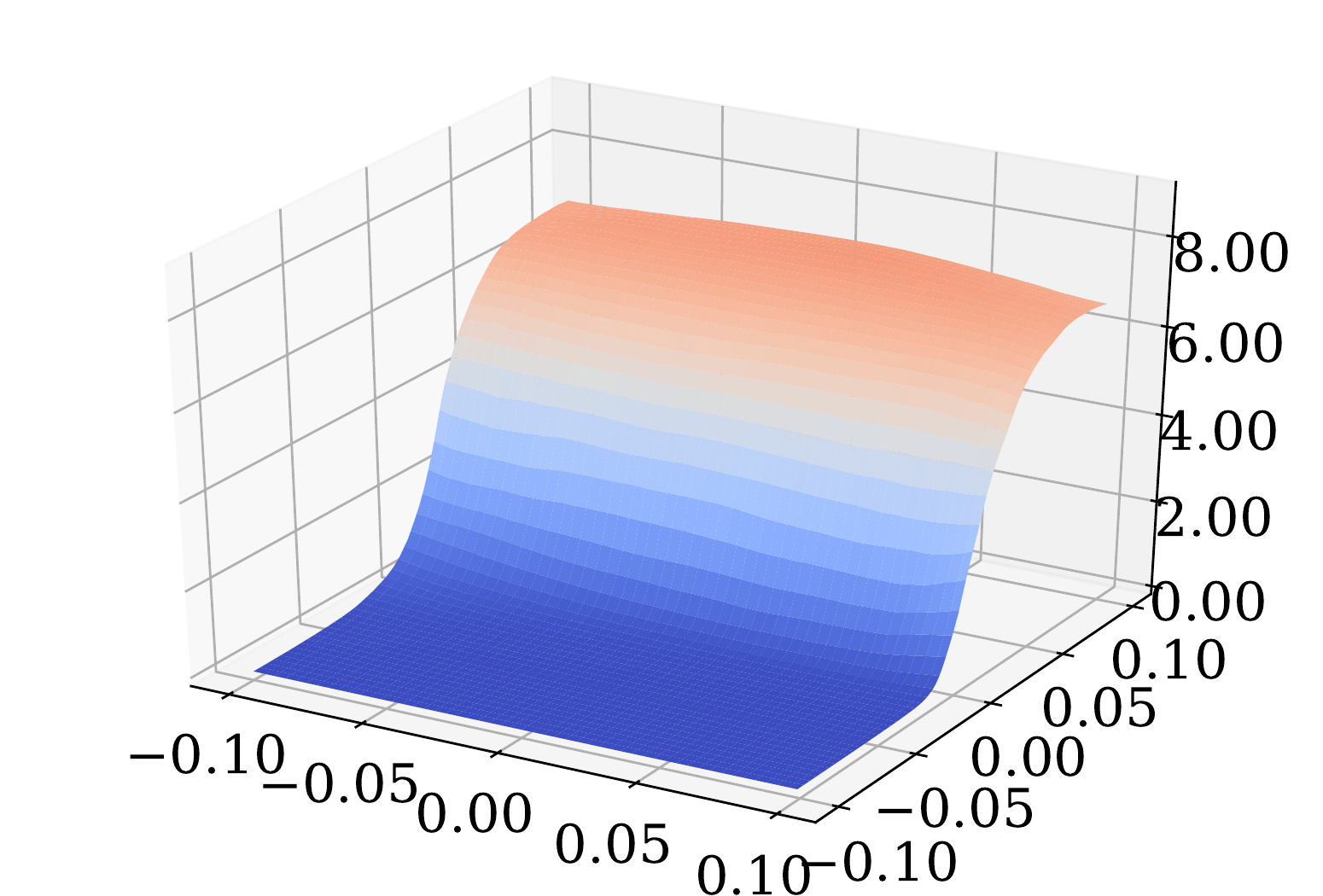}}
		&
		\subfloat[CAT]{\includegraphics[width=0.225\textwidth]{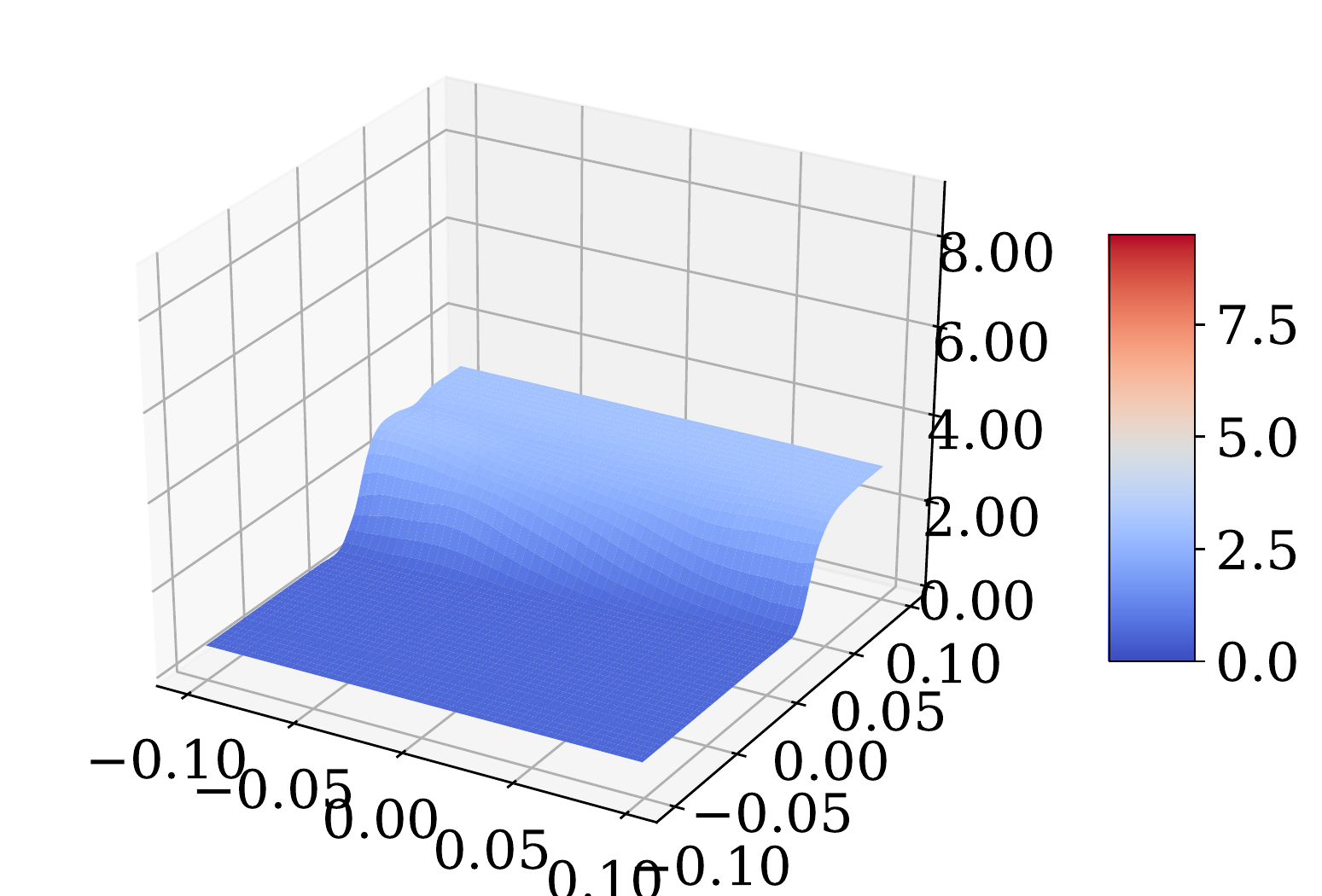}}
	\end{tabular}
		\caption{Loss landscape comparison of different adversarial training methods}
		\label{fig:loss_landscape}
\end{figure*}  

\subsection{Experimental Setup}

\textbf{Dataset and model structure.} We use two popular dataset CIFAR-10~\citep{cifar10} and Restricted-ImageNet~\citep{deng2009imagenet} 
for performance evaluation. For CIFAR-10, we use both standard VGG-16~\citep{simonyan2014very} and Wide ResNet that is used in both vanilla adversarial training \citep{madry2017towards} and TRADES \citep{zhang2019theoretically}. 
For VGG-16, we implement adversarial training with the standard hyper-parameters and train TRADES with the official implementation.
For Wide ResNet, since the model has become standard for testing adversarial training methods, we use exactly the same model structure provided by \cite{madry2017towards,zhang2019theoretically}.
 And use the models' checkpoint released by adversarial training and TRADES official repository and implement the Madry's adversarial training using the standard hyper-parameters. For Restricted-ImageNet, we use ResNet-50.
 All our experiments were implemented in Pytorch-1.4  and conducted using dual Intel E5-2640 v4 CPUs (2.40GHz) with 512 GB memory with a GTX 2080 TI GPU. 

\textbf{Implementation details.} 
We set the number of iterations in adversarial attack to be 10 for all methods. Adversarial training and TRADES are trained on PGD attacks setting $\epsilon=8/255$ with cross entropy loss (CE). We implement our CAT method both on cross entropy (CE) \cite{madry2017towards} and C$\&$W loss \cite{carlini2017towards}, and set $\epsilon_{\max} =8/255$. All the models are trained using SGD with momentum
0.9, weight decay $5\times 10^{-4}$. For VGG-16/Wide ResNet models, we use the initial learning rate of 0.01/0.1, and we decay the learning rate by 90\% at the 80th, 140th, and 180th epoch. 
For CAT, we set epsilon scheduling parameter $\eta=0.005$, $\epsilon_{max}=8/255$ and weighting parameter $c=10$. For CAT-MIX, we set $\kappa=10$.

\subsection{Robustness Evaluation and Analysis}
\textbf{White-box attcks.} For CIFAR10, we evaluate all the models under different tolerance of white-box $\ell_\infty$-norm bounded non-targeted PGD and C$\&$W attack. Specifically, we use both $\text{PGD}^{20}$ (20-step PGD with step size $\epsilon/5)$ and $\text{C\&W}_\infty$. All attacks are equipped with random-start. To be noted, when $\epsilon=0$, the robust accuracy is reduced to test accuracy of unperturbed (natural) test samples, i.e clean accuracy.

The experimental results are shown in Figure \ref{fig:res}, where we can easily see that
both CAT-CE and CAT-MIX clearly outperform other methods among $\epsilon$ from $0$ to $0.07$. So our methods can achieve better robust error at the standard $8/255$ perturbation threshold considered in the literature, and also has better clean accuracy ($\epsilon=0$).
The accuracy curve becomes quite flat when $\epsilon$ is increased. 


Wide ResNet has become a standard structure for comparing adversarial training methods, and it's standard to train and evaluate with $8/255$ $\ell_\infty$ norm perturbation. 
For this setting, 
we collect the reported accuracy from 7 other adversarial training methods, with several of them published very recently, to have a detailed full comparison.
As shown in Table \ref{tb:main},  our method achieves state-of-art robust accuracy while maintaining a high clean accuracy. 
Due to the page limit, we put the Restricted ImageNet result in the appendix. 

\textbf{Black-box transfer attacks.} 
We follow the criterion of evaluating transfer attacks as suggested by \citet{athalye2018obfuscated} to inspect whether the models trained by CAT will cause the issue of obfuscated gradients and give a false sense of model robustness.
We generate 10,000 adversarial examples of CIFAR-10 from natural models with $\epsilon=0.03$ and evaluate their attack performance on the target model. 
Table \ref{tab:transfer} shows that CAT achieves the best accuracy compared with adversarial training and TRADES, suggesting the effectiveness of CAT in defending both white-box and transfer attacks. 

\begin{table}[htbp]
    \caption{Robust accuracy under transfer attack on CIFAR-10
    }
    \centering
    \begin{tabular}{c|c|c}
        \toprule
        Method &VGG 16& Wide ResNet \\
       \hline
       Adv train & 79.13\% & 85.84\% \\
       TRADES & 83.53\%& 83.90\% \\
       CAT & {\bf 86.58\%}& {\bf 88.66 \%}\\
       \bottomrule
    \end{tabular}
    \label{tab:transfer}
\end{table}

\subsection{Loss Landscape Exploration}
To further verify the superior robustness using CAT,
we visualize the loss landscape of different training methods in Figure \ref{fig:loss_landscape}. Following the implementation in \citep{engstrom2018evaluating}, we divide the data input along a linear space grid defined by the sign of the input gradient and a random Rademacher vector, where the x- and y- axes represent the magnitude of the perturbation added in each direction and the z-axis represents the loss. 

As shown in Figure \ref{fig:loss_landscape}, CAT generates a model with a lower and smoother loss landscape. Also, it could be taken as another strong evidence that we have found a robust model through CAT training.

\subsection{Ablation study}
\label{sec:aba}

\begin{figure}[tb]
    \centering
    \begin{tabular}{c}
 \subfloat[\label{fig:num_aba}]{\includegraphics[width=0.4\textwidth]{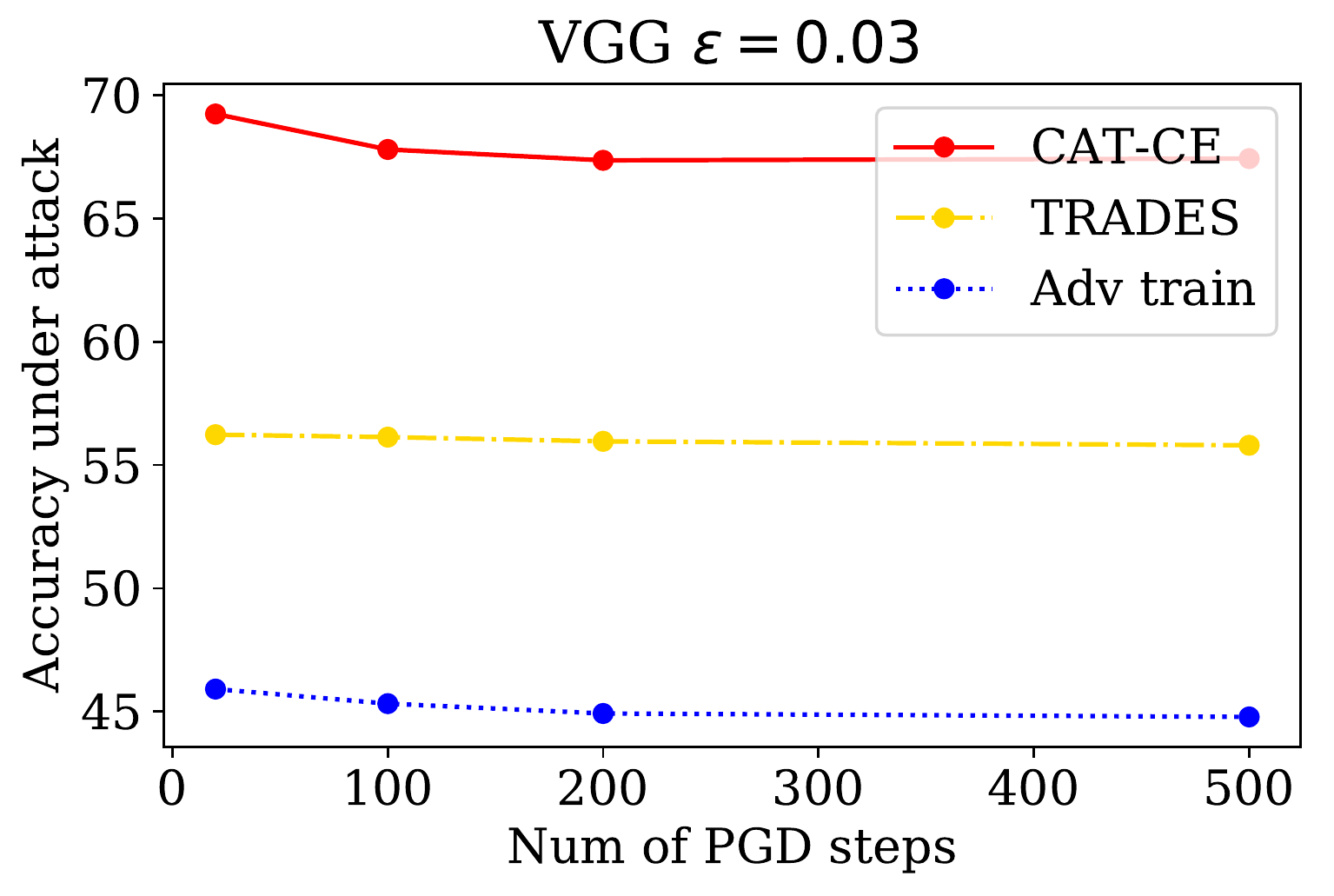}}
		\\
		\subfloat[\label{fig:label}]{\includegraphics[width=0.4\textwidth]{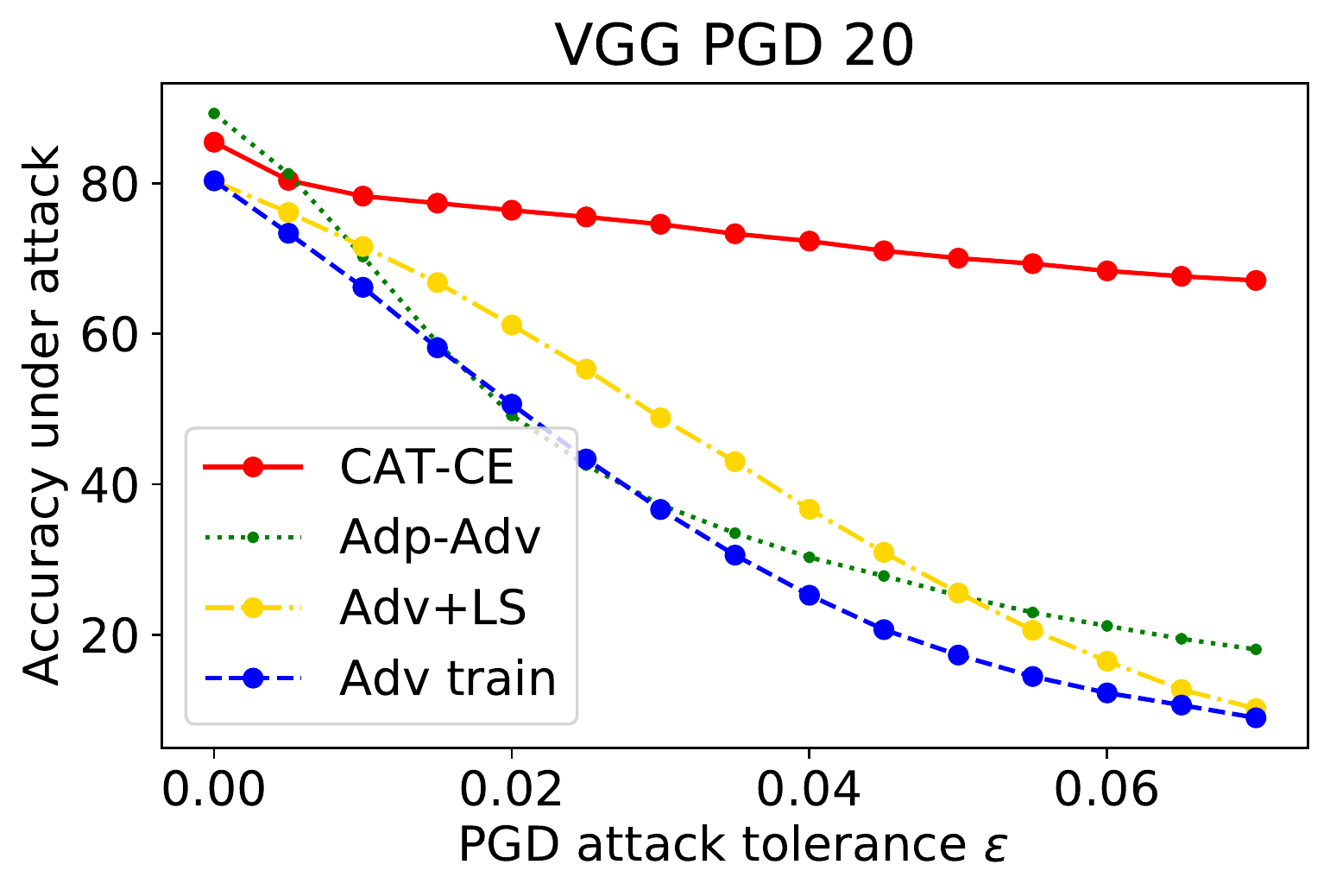}}
        \\ 
	\end{tabular}
		\caption{Analysis of CAT. In (a) we test CAT under different steps of PGD attack, and in (b) we conduct an ablation study on CAT by changing the loss function and removing Label Adaption (LA). }
	    
\end{figure}

\textbf{The importance of adaptive label uncertainty.}
Here we discuss and perform an ablation study using VGG-16 and CIFAR-10 on the importance of adaptive label uncertainty and adaptive instance-wise $\epsilon$. 
In Figure \ref{fig:label}, Adp train denotes the original adversarial training, Adv+LS denotes adversarial training with label smoothing (setting $y$ by Eq~\eqref{eq:tildey_new}), Adp-Adv denotes adversarial training with adaptive instance-wise $\epsilon$, and CAT-CE is the proposed method which is a combination of these two tricks. We found that only applying adaptive instance-wise $\epsilon$  or label smoothing cannot significantly boost the robust accuracy over standard adversarial training, but the proposed method, by nicely  combining these two ideas, can significantly improve the performance. 
This explains why CAT significantly outperforms some instance adaptive $\epsilon$ methods like IAAT and MMA. 

\textbf{More iterations of PGD attack.} As suggested in \citet{athalye2018obfuscated}, to verify that the performance gain is not brought by insufficient iterations in PGD attack, in Figure \ref{fig:num_aba}, we show the robust accuracy with the number of iteration varying from $10$ to $500$ on CIFAR10 VGG16. The results show that although increasing the number of iterations would decrease the performance by around 2\%, CAT  always outperforms other methods significantly.

\section{Conclusions}
In this paper, we propose CAT, a customized adversarial training method that is designed to have better generalization for both clean and robust performance. 
We also provide a theoretical analysis to motivate our algorithm. 
Experimental results show that CAT has achieved state-of-art robust accuracy and a high clean accuracy while keeping similar running time as standard adversarial training. The success of CAT indicates that it is crucial to customize the perturbation level on both data sample side and its label in adversarial training. 

\bibliography{icml2020}

\begin{thebibliography}{39}
\providecommand{\natexlab}[1]{#1}
\providecommand{\url}[1]{\texttt{#1}}
\expandafter\ifx\csname urlstyle\endcsname\relax
  \providecommand{\doi}[1]{doi: #1}\else
  \providecommand{\doi}{doi: \begingroup \urlstyle{rm}\Url}\fi

\bibitem[Athalye et~al.(2018)Athalye, Carlini, and
  Wagner]{athalye2018obfuscated}
Athalye, A., Carlini, N., and Wagner, D.
\newblock Obfuscated gradients give a false sense of security: Circumventing
  defenses to adversarial examples.
\newblock \emph{International Coference on International Conference on Machine
  Learning}, 2018.

\bibitem[Balaji et~al.(2019)Balaji, Goldstein, and Hoffman]{balaji2019instance}
Balaji, Y., Goldstein, T., and Hoffman, J.
\newblock Instance adaptive adversarial training: Improved accuracy tradeoffs
  in neural nets.
\newblock \emph{arXiv preprint arXiv:1910.08051}, 2019.

\bibitem[Brendel et~al.(2017)Brendel, Rauber, and Bethge]{brendel2017decision}
Brendel, W., Rauber, J., and Bethge, M.
\newblock Decision-based adversarial attacks: Reliable attacks against
  black-box machine learning models.
\newblock \emph{arXiv preprint arXiv:1712.04248}, 2017.

\bibitem[Carlini \& Wagner(2017)Carlini and Wagner]{carlini2017towards}
Carlini, N. and Wagner, D.
\newblock Towards evaluating the robustness of neural networks.
\newblock In \emph{IEEE Symposium on Security and Privacy}, pp.\  39--57, 2017.

\bibitem[Chen et~al.(2017)Chen, Zhang, Sharma, Yi, and Hsieh]{chen2017zoo}
Chen, P.-Y., Zhang, H., Sharma, Y., Yi, J., and Hsieh, C.-J.
\newblock Zoo: Zeroth order optimization based black-box attacks to deep neural
  networks without training substitute models.
\newblock In \emph{Proceedings of the 10th ACM Workshop on Artificial
  Intelligence and Security}, pp.\  15--26, 2017.

\bibitem[Cheng et~al.(2018)Cheng, Le, Chen, Yi, Zhang, and
  Hsieh]{cheng2018query}
Cheng, M., Le, T., Chen, P.-Y., Yi, J., Zhang, H., and Hsieh, C.-J.
\newblock Query-efficient hard-label black-box attack: An optimization-based
  approach.
\newblock \emph{arXiv preprint arXiv:1807.04457}, 2018.

\bibitem[Cheng et~al.(2020)Cheng, Singh, Chen, Chen, Liu, and
  Hsieh]{cheng2020signopt}
Cheng, M., Singh, S., Chen, P., Chen, P.-Y., Liu, S., and Hsieh, C.-J.
\newblock Sign-opt: A query-efficient hard-label adversarial attack.
\newblock In \emph{ICLR}, 2020.

\bibitem[Cohen et~al.(2019)Cohen, Rosenfeld, and Kolter]{cohen2019certified}
Cohen, J.~M., Rosenfeld, E., and Kolter, J.~Z.
\newblock Certified adversarial robustness via randomized smoothing.
\newblock \emph{International Conference on Machine Learning}, 2019.

\bibitem[Deng et~al.(2009)Deng, Dong, Socher, Li, Li, and
  Fei-Fei]{deng2009imagenet}
Deng, J., Dong, W., Socher, R., Li, L.-J., Li, K., and Fei-Fei, L.
\newblock Imagenet: A large-scale hierarchical image database.
\newblock In \emph{IEEE Conference on Computer Vision and Pattern Recognition},
  pp.\  248--255, 2009.

\bibitem[Ding et~al.(2018)Ding, Sharma, Lui, and Huang]{ding2018max}
Ding, G.~W., Sharma, Y., Lui, K. Y.~C., and Huang, R.
\newblock Max-margin adversarial (mma) training: Direct input space margin
  maximization through adversarial training.
\newblock \emph{arXiv preprint arXiv:1812.02637}, 2018.

\bibitem[Engstrom et~al.(2018)Engstrom, Ilyas, and
  Athalye]{engstrom2018evaluating}
Engstrom, L., Ilyas, A., and Athalye, A.
\newblock Evaluating and understanding the robustness of adversarial logit
  pairing.
\newblock \emph{arXiv preprint arXiv:1807.10272}, 2018.

\bibitem[Eykholt et~al.(2018)Eykholt, Evtimov, Fernandes, Li, Rahmati, Xiao,
  Prakash, Kohno, and Song]{eykholt2018robust}
Eykholt, K., Evtimov, I., Fernandes, E., Li, B., Rahmati, A., Xiao, C.,
  Prakash, A., Kohno, T., and Song, D.
\newblock Robust physical-world attacks on deep learning visual classification.
\newblock In \emph{Proceedings of the IEEE Conference on Computer Vision and
  Pattern Recognition}, pp.\  1625--1634, 2018.

\bibitem[Gao et~al.(2019)Gao, Cai, Li, Hsieh, Wang, and
  Lee]{gao2019convergence}
Gao, R., Cai, T., Li, H., Hsieh, C.-J., Wang, L., and Lee, J.~D.
\newblock Convergence of adversarial training in overparametrized neural
  networks.
\newblock In \emph{Advances in Neural Information Processing Systems}, pp.\
  13009--13020, 2019.

\bibitem[Goibert \& Dohmatob(2019)Goibert and Dohmatob]{goibert2019adversarial}
Goibert, M. and Dohmatob, E.
\newblock Adversarial robustness via adversarial label-smoothing.
\newblock \emph{arXiv preprint arXiv:1906.11567}, 2019.

\bibitem[Goodfellow et~al.(2015)Goodfellow, Shlens, and
  Szegedy]{goodfellow2014explaining}
Goodfellow, I.~J., Shlens, J., and Szegedy, C.
\newblock Explaining and harnessing adversarial examples.
\newblock \emph{International Conference on Learning Representations}, 2015.

\bibitem[Ilyas et~al.(2018)Ilyas, Engstrom, Athalye, and Lin]{ilyas2018black}
Ilyas, A., Engstrom, L., Athalye, A., and Lin, J.
\newblock Black-box adversarial attacks with limited queries and information.
\newblock \emph{arXiv preprint arXiv:1804.08598}, 2018.

\bibitem[Krizhevsky et~al.()Krizhevsky, Nair, and Hinton]{cifar10}
Krizhevsky, A., Nair, V., and Hinton, G.
\newblock Cifar-10 (canadian institute for advanced research).
\newblock URL \url{http://www.cs.toronto.edu/~kriz/cifar.html}.

\bibitem[Kurakin et~al.(2017)Kurakin, Goodfellow, and
  Bengio]{kurakin2016adversarial_ICLR}
Kurakin, A., Goodfellow, I., and Bengio, S.
\newblock Adversarial machine learning at scale.
\newblock \emph{International Conference on Learning Representations}, 2017.

\bibitem[Liu et~al.(2018)Liu, Cheng, Zhang, and Hsieh]{liu2018towards}
Liu, X., Cheng, M., Zhang, H., and Hsieh, C.-J.
\newblock Towards robust neural networks via random self-ensemble.
\newblock In \emph{Proceedings of the European Conference on Computer Vision
  (ECCV)}, pp.\  369--385, 2018.

\bibitem[Madry et~al.(2018)Madry, Makelov, Schmidt, Tsipras, and
  Vladu]{madry2017towards}
Madry, A., Makelov, A., Schmidt, L., Tsipras, D., and Vladu, A.
\newblock Towards deep learning models resistant to adversarial attacks.
\newblock \emph{International Conference on Learning Representations}, 2018.

\bibitem[Shafahi et~al.(2018)Shafahi, Ghiasi, Huang, and
  Goldstein]{shafahi2018label}
Shafahi, A., Ghiasi, A., Huang, F., and Goldstein, T.
\newblock Label smoothing and logit squeezing: A replacement for adversarial
  training?
\newblock 2018.

\bibitem[Shafahi et~al.(2019)Shafahi, Najibi, Ghiasi, Xu, Dickerson, Studer,
  Davis, Taylor, and Goldstein]{shafahi2019adversarial}
Shafahi, A., Najibi, M., Ghiasi, A., Xu, Z., Dickerson, J., Studer, C., Davis,
  L.~S., Taylor, G., and Goldstein, T.
\newblock Adversarial training for free!
\newblock \emph{Neural Information Processing Systems}, 2019.

\bibitem[Simonyan \& Zisserman(2015)Simonyan and Zisserman]{simonyan2014very}
Simonyan, K. and Zisserman, A.
\newblock Very deep convolutional networks for large-scale image recognition.
\newblock \emph{International Conference on Learning Representations}, 2015.

\bibitem[Su et~al.(2018)Su, Zhang, Chen, Yi, Chen, and Gao]{su2018robustness}
Su, D., Zhang, H., Chen, H., Yi, J., Chen, P.-Y., and Gao, Y.
\newblock Is robustness the cost of accuracy?--a comprehensive study on the
  robustness of 18 deep image classification models.
\newblock In \emph{Proceedings of the European Conference on Computer Vision
  (ECCV)}, pp.\  631--648, 2018.

\bibitem[Szegedy et~al.(2014)Szegedy, Zaremba, Sutskever, Bruna, Erhan,
  Goodfellow, and Fergus]{szegedy2013intriguing}
Szegedy, C., Zaremba, W., Sutskever, I., Bruna, J., Erhan, D., Goodfellow, I.,
  and Fergus, R.
\newblock Intriguing properties of neural networks.
\newblock \emph{International Conference on Learning Representations}, 2014.

\bibitem[Szegedy et~al.(2016)Szegedy, Vanhoucke, Ioffe, Shlens, and
  Wojna]{szegedy2016rethinking}
Szegedy, C., Vanhoucke, V., Ioffe, S., Shlens, J., and Wojna, Z.
\newblock Rethinking the inception architecture for computer vision.
\newblock In \emph{Proceedings of the IEEE conference on computer vision and
  pattern recognition}, pp.\  2818--2826, 2016.

\bibitem[Thulasidasan et~al.(2019)Thulasidasan, Chennupati, Bilmes,
  Bhattacharya, and Michalak]{thulasidasan2019mixup}
Thulasidasan, S., Chennupati, G., Bilmes, J., Bhattacharya, T., and Michalak,
  S.
\newblock On mixup training: Improved calibration and predictive uncertainty
  for deep neural networks.
\newblock \emph{Neural Information Processing Systems}, 2019.

\bibitem[Tsipras et~al.(2019)Tsipras, Santurkar, Engstrom, Turner, and
  Madry]{tsipras2019robustness}
Tsipras, D., Santurkar, S., Engstrom, L., Turner, A., and Madry, A.
\newblock Robustness may be at odds with accuracy.
\newblock In \emph{International Conference on Learning Representations}, 2019.

\bibitem[Verma et~al.(2018)Verma, Lamb, Beckham, Najafi, Mitliagkas, Courville,
  Lopez-Paz, and Bengio]{verma2018manifold}
Verma, V., Lamb, A., Beckham, C., Najafi, A., Mitliagkas, I., Courville, A.,
  Lopez-Paz, D., and Bengio, Y.
\newblock Manifold mixup: Better representations by interpolating hidden
  states.
\newblock \emph{International Conference on Machine Learning}, 2018.

\bibitem[Wang(2019)]{wang2018bilateral}
Wang, J.
\newblock Bilateral adversarial training: Towards fast training of more robust
  models against adversarial attacks.
\newblock \emph{International Conference on Computer Vision}, 2019.

\bibitem[Wang et~al.(2019)Wang, Ma, Bailey, Yi, Zhou, and
  Gu]{wang2019convergence}
Wang, Y., Ma, X., Bailey, J., Yi, J., Zhou, B., and Gu, Q.
\newblock On the convergence and robustness of adversarial training.
\newblock In \emph{International Conference on Machine Learning}, pp.\
  6586--6595, 2019.

\bibitem[Wang(2020)]{zou2018improving}
Wang, Zou, Y. B. M.~G.
\newblock Improving adversarial robustness requires revisiting misclassified
  examples.
\newblock \emph{International Conference on Learning Representations}, 2020.
\newblock URL \url{https://openreview.net/forum?id=rklOg6EFwS}.

\bibitem[Wei \& Ma(2019)Wei and Ma]{wei2019improved}
Wei, C. and Ma, T.
\newblock Improved sample complexities for deep networks and robust
  classification via an all-layer margin.
\newblock \emph{arXiv preprint arXiv:1910.04284}, 2019.

\bibitem[Wong et~al.(2020)Wong, Rice, and Kolter]{wong2020fast}
Wong, E., Rice, L., and Kolter, J.~Z.
\newblock Fast is better than free: Revisiting adversarial training.
\newblock \emph{arXiv preprint arXiv:2001.03994}, 2020.

\bibitem[Yin et~al.(2018)Yin, Ramchandran, and Bartlett]{yin2018rademacher}
Yin, D., Ramchandran, K., and Bartlett, P.
\newblock Rademacher complexity for adversarially robust generalization.
\newblock \emph{arXiv preprint arXiv:1810.11914}, 2018.

\bibitem[Zantedeschi et~al.(2017)Zantedeschi, Nicolae, and
  Rawat]{zantedeschi2017efficient}
Zantedeschi, V., Nicolae, M.-I., and Rawat, A.
\newblock Efficient defenses against adversarial attacks.
\newblock In \emph{ACM Workshop on Artificial Intelligence and Security}, pp.\
  39--49, 2017.

\bibitem[Zhang et~al.(2019{\natexlab{a}})Zhang, Zhang, Lu, Zhu, and
  Dong]{zhang2019you}
Zhang, D., Zhang, T., Lu, Y., Zhu, Z., and Dong, B.
\newblock You only propagate once: Accelerating adversarial training via
  maximal principle.
\newblock In \emph{Advances in Neural Information Processing Systems}, pp.\
  227--238, 2019{\natexlab{a}}.

\bibitem[Zhang et~al.(2018)Zhang, Cisse, Dauphin, and
  Lopez-Paz]{zhang2017mixup}
Zhang, H., Cisse, M., Dauphin, Y.~N., and Lopez-Paz, D.
\newblock mixup: Beyond empirical risk minimization.
\newblock \emph{International Conference on Learning Representations}, 2018.

\bibitem[Zhang et~al.(2019{\natexlab{b}})Zhang, Yu, Jiao, Xing, Ghaoui, and
  Jordan]{zhang2019theoretically}
Zhang, H., Yu, Y., Jiao, J., Xing, E.~P., Ghaoui, L.~E., and Jordan, M.~I.
\newblock Theoretically principled trade-off between robustness and accuracy.
\newblock \emph{International Conference on Machine Learning},
  2019{\natexlab{b}}.

\end{thebibliography}
\bibliographystyle{icml2020}
    
\appendix
\onecolumn

\section{Omitted Proofs}
In this section we provide the omitted proof for Theorem \ref{thm:main}, which is adapted from Theorem 2.1 from \cite{wei2019improved}. They defined the all layer margin for a $k$-layer network $h_{\theta}(\bx)=f_{k}\circ f_{k-1}\circ \cdots f_1(\bx)$ and perturbation $\delta=(\bdelta_1,\bdelta_2,\cdots \bdelta_k)$ as follows:
\begin{align*}
    h_1(\bx,\delta) = & f_1(\bx) + \bdelta_1\|\bx\|_2\\
    h_i(\bx,\delta)= &f_i(h_{i-1}(\bx,\bdelta))+\bdelta_i\|h_{i-1}(\bx,\delta)\|_2\\
    H_\theta(\bx,\delta) = & h_k(\bx,\delta). 
\end{align*} 
They define the all-layer margin as the minimum norm of $\delta=(\bdelta_i)_{i=1}^k$ required that causes the classifier to make a false prediction. 
\begin{equation}
    \begin{aligned}
    \label{eqn:margin}
    m_F(\bx,y):=& \min_{\bdelta^i, \bdelta^o} \sqrt{ \|\bdelta^i\|^2+\|\bdelta^o\|^2}\\
    &\text{ subject to } \max_{y'} H_\theta(\bx,\bdelta^i,\bdelta^o)_{y'}\neq y. 
    \end{aligned} 
\end{equation}
They consider the function class $\cF = \{f_k\circ f_{k-1}\cdots \circ f_1:f_i\in \cF_i \}$ be the class of compositions of functions from function classes $\cF_1,\cdots \cF_k$.  
 They achieve the generalization bound as follows:
\begin{theorem}[Theorem 2.1 from \cite{wei2019improved}]
\label{thm:cited}
In the above setting, with probability $1-\delta$ over the draw of the data, all classifiers $F\in \cF$ which achieve training error $0$ satisfy
\begin{equation*}
    \mathbb{E}[f_\theta(\bx)=y] \lesssim \frac{\sum_i\cC_i \log^2 n}{\sqrt{n}}\sqrt{\frac{1}{n}\sum_{i=1}^n \frac{1}{m_F(\bx_i,y_i)} } + \zeta,
\end{equation*}
where $\zeta$ is of small order $O\left(\frac{1}{n}\log(1/\delta)\right)$.
\end{theorem}
For our problem, we define $h_\theta(\bx):=f_2\circ f_1(\bx)$, where $f_1$ is identity mapping, and $f_2$ is the original function $h_\theta$. Therefore the all layer margin is reduced to our bilateral margin:
\begin{align*}
h_1(\bx,\bdelta^{i}) = & f_1(\bx)+\bdelta^i\|\bx\|_2 =  \bx+\bdelta^i\|\bx\|\\
H_\theta(\bx,\bdelta)= h_2(\bx,\bdelta^i,\bdelta^o)=& f_2(h_1(\bx,\bdelta^i))+\bdelta^o\|h_1(\bx,\bdelta^i)\|\\= & h_\theta(\bx+\bdelta^i\|\bx\|)+\bdelta^o\|\bx+\bdelta^{i}\|\bx\|\|.     
\end{align*}
Next, notice since $f_1$ is identity mapping, and composition with $h_\theta$ doesn't affect the overall complexity. We apply Theorem \ref{thm:cited} and get our result.

\end{document}